\documentclass{article}
\PassOptionsToPackage{numbers, compress}{natbib}
\PassOptionsToPackage{table}{xcolor}
\usepackage[main, final]{neurips_2026}

\usepackage[utf8]{inputenc}
\usepackage[T1]{fontenc}
\usepackage{amsmath}
\usepackage{amssymb}
\usepackage{amsfonts}
\usepackage{amsthm}
\usepackage{mathtools}
\usepackage{bm}
\usepackage{nicefrac}
\usepackage{xspace}
\usepackage{xcolor}
\usepackage{graphicx}
\usepackage{booktabs}
\usepackage{multirow}
\usepackage{makecell}
\usepackage{array}
\usepackage{tabularx}
\usepackage{threeparttable}
\usepackage{caption}
\usepackage{float}
\usepackage{algorithm}
\usepackage{algpseudocode}
\usepackage{enumitem}
\usepackage{marvosym}
\usepackage{microtype}
\usepackage{url}
\usepackage{hyperref}
\newcommand{\ViReIntroFigureWidth}{0.97\textwidth}
\newcommand{\ViReMethodFigureWidth}{0.88\textwidth}
\newcommand{\ViReAfterIntroFigureSpace}{0mm}
\newcommand{\ViReAfterAbstractSpace}{0mm}
\newcommand{\ViReBibSep}{4.15pt plus 0.3pt minus 0.2pt}
\newcommand{\ViReAugmentationStretch}{1.20}
\newcommand{\ViReImplementationStretch}{1.14}
\newcommand{\ViReFullResultsWidth}{1.00\textwidth}
\newcommand{\ViReFullResultsStretch}{1.20}

\hypersetup{colorlinks=true,linkcolor=blue,citecolor=blue,urlcolor=blue,
  pdftitle={Revitalizing Medical Time Series with Vision-Informed Retrieval: A Vision-Language Perspective},
  pdfauthor={Guoqi Yu, Juncheng Wang, Shujun Wang},
  pdfsubject={NeurIPS 2026},
  pdfkeywords={medical time series, EEG, ECG, vision-language models, CLIP, cross-modal retrieval}}
\newcommand{\redref}[1]{\begingroup\hypersetup{linkcolor=red}\ref{#1}\endgroup}
\newcommand{\figref}[1]{Figure~\redref{#1}}
\newcommand{\tabref}[1]{Table~\redref{#1}}
\providecommand{\algref}{}
\renewcommand{\algref}[1]{Algorithm~\redref{#1}}
\newcommand{\appref}[1]{Appendix~\ref{#1}}

\microtypesetup{protrusion=true,expansion=true}
\ifdefined\pdfimageresolution\pdfimageresolution=300\fi

\newcommand{\eqparbreak}{\par\penalty0\relax}

\theoremstyle{plain}

\theoremstyle{definition}

\theoremstyle{remark}

\definecolor{purple}{RGB}{235,222,240}
\definecolor{softband}{RGB}{247,244,249}
\definecolor{revmaroon}{RGB}{128,0,0}
\newcommand{\first}[1]{\textbf{\textcolor{revmaroon}{#1}}}
\newcommand{\second}[1]{\textcolor{blue}{\textit{#1}}}
\newcommand{\std}[1]{\scriptsize{$\pm$#1}}
\newcommand{\define}[1]{\vspace{0mm}\noindent\textbf{#1.}}

\newcommand{\ViReRepoURL}{https://github.com/Levi-Ackman/ViRe}
\newcommand{\ViReNotebookURL}{https://github.com/Levi-Ackman/ViRe}

\title{Revitalizing Medical Time Series with Vision-Informed Retrieval: A Vision-Language Perspective}

\author{%
  Guoqi Yu \qquad Juncheng Wang \qquad Shujun Wang\textsuperscript{\Letter} \\[1pt]
  Department of Biomedical Engineering and Sports Technology \\[1pt]
  The Hong Kong Polytechnic University \\[1pt]
  \textsuperscript{\Letter}\,Correspondence to: Shujun Wang (e-mail: \texttt{shu-jun.wang@polyu.edu.hk})
}

\begin{document}
\maketitle

\begin{center}
  \captionsetup{type=figure}
  \includegraphics[width=\ViReIntroFigureWidth]{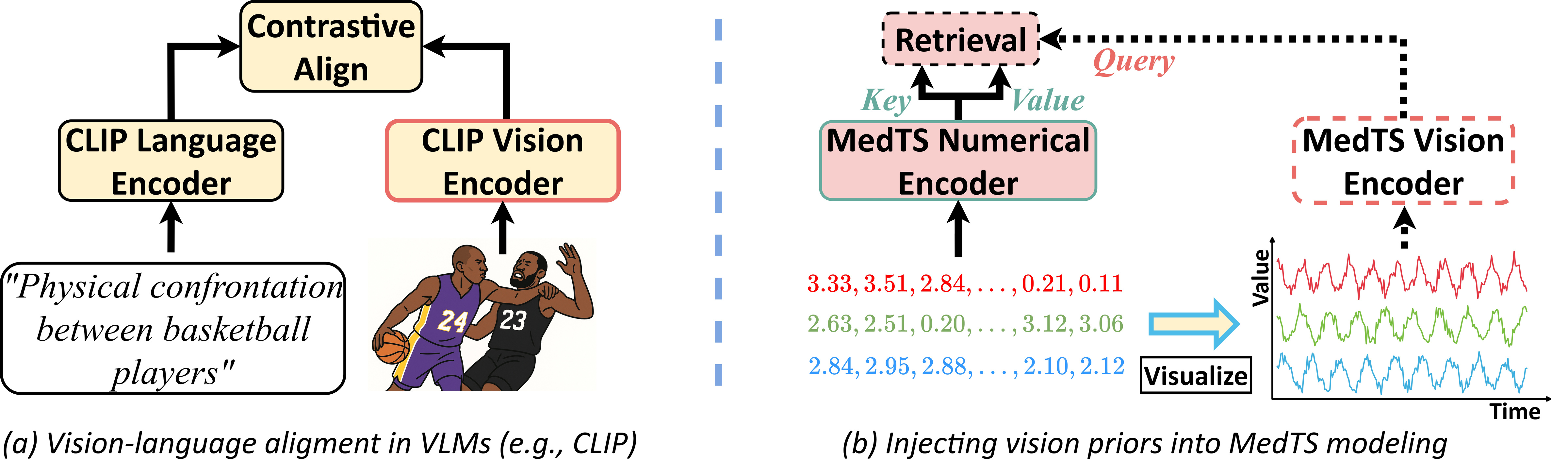}
  \vspace{-2mm}
  \captionof{figure}{Existing Medical Time Series (MedTS) models operate primarily on numerical sequences, overlooking the waveform morphology that clinicians rely on. We propose to use a frozen CLIP Vision Encoder to turn a deterministic waveform rendering into a morphology-aware \textit{Vision Query}, which retrieves relevant temporal and channel evidence from numerical MedTS features.}
  \label{fig:intro}
\end{center}
\vspace{\ViReAfterIntroFigureSpace}

\begin{abstract}
Medical time series (MedTS) underpin many clinical classification tasks, yet existing methods usually represent them only as numerical sequences and underuse the morphology that is explicit in waveform inspection. To bridge this gap, we introduce Vision-Informed Retrieval (ViRe), which uses a frozen VLM-derived waveform representation as a morphology-aware Query to guide retrieval from raw numerical MedTS features. Specifically, a Vision Query is extracted using pre-trained vision-language models (VLMs) to obtain morphology-aware priors from waveform plots. A tailored attention-based cross-modal retrieval mechanism then uses the Vision Query to select morphology-relevant temporal and channel evidence from the numerical representation. ViRe demonstrates strong effectiveness against ten established baselines, yielding an overall \textbf{6.42\%} relative improvement over the previous state of the art across six public benchmarks. Code, training scripts, and reproducibility materials are publicly available in the \href{\ViReRepoURL}{\textbf{GitHub Repo}}.
\end{abstract}

\vspace{\ViReAfterAbstractSpace}
\section{Introduction}
Medical time series (MedTS), such as electrocardiography (ECG)~\citep{alghatrif2012centralecg} and electroencephalography (EEG)~\citep{cohen2017eeg1}, provide continuous records of heart and brain activity, forming the cornerstone of modern diagnostics. Many routine clinical tasks, including epilepsy detection~\citep{ullah2018epilepsy}, sleep staging~\citep{jiao2020eog0}, arrhythmia screening~\citep{Jin2024ecg1}, and cardiovascular risk stratification~\citep{zhou2023semi}, are generally cast as MedTS classification. These tasks depend on both localized temporal events and relationships among channels.

The advent of artificial intelligence enables fast, automated classification, greatly improving diagnostic efficiency. Early machine learning relied on extracting handcrafted features (\textit{e.g.}, band power, Hjorth parameters) from numerical input~\citep{tzimourta2021mlmed,al2023mlmed}. Deep learning shifted this traditional paradigm, enabling convolutional and recurrent networks to predict directly from raw numerical data~\citep{arif2024ef,tang2021gnn}. More recently, Transformer models further improved performance by capturing longer temporal dependencies~\citep{wang2024medformer}. Across these architectures, however, each record remains a numerical sequence or matrix. Yet the same recordings are also inspected as waveforms by clinicians, where local shape and cross-channel structure are directly visible to the trained eye.

Despite the architectural advances, prevailing models adopt a reductive premise: \emph{they treat MedTS purely as numerical sequences and underuse the morphology-centric structure that is explicit in waveform inspection}. In real clinical workflows, however, raw EEG and ECG recordings are visually reviewed by neurologists or cardiologists~\citep{nejedly2019eegvision}. They summarize waveform morphology (\textit{e.g.}, P--QRS--T complexes, ST-segment deviations, epileptiform discharges, and cross-channel co-activations) into structured reports and quantitative indices for referring clinicians~\citep{da1999eegvision,kligfield2007ecgvision}. Although clinicians primarily consume these reports rather than raw waveforms, visual waveform patterns remain crucial for establishing the diagnostic criteria of MedTS~\citep{kural2020eegvision,tatum2012eegvision}. By focusing solely on raw numerical matrices, current MedTS models largely overlook this morphology-centric perspective and underutilize the strong inductive structure tied to waveform shape, making it harder to reach diagnoses that align with expert reasoning~\citep{wang2024medformer}. \emph{Bridging this mismatch between numerics-oriented representations and waveform-based clinical reasoning is therefore critical for building MedTS models that are accurate, reliable, and clinically well-grounded.}

This mismatch points to an opportunity: since vision representations are essential for diagnosis, MedTS models should also reason over waveform visualizations, rather than rely entirely on numerical signals. However, training models that can extract such waveform representations aligned with human concepts from scratch is data-hungry and brittle under subject/device shifts~\citep{nguyen2025gen2,mehari2022gen7}. A natural remedy is to leverage an existing visual representation extractor that has already been trained to align image structure with human knowledge, and use it to provide compact, morphology-aware priors. This makes \textbf{Vision Language Models} (VLMs) (\textit{e.g.}, CLIP~\citep{radford2021clip}) particularly attractive to our problem. Trained with natural-language supervision, their vision encoders are optimized to organize images in a space that is aligned with \textbf{human-describable concepts}~\citep{gong2025clip,bhalla2024clip}. By feeding waveforms into these encoders, the resulting embeddings provide a semantically structured summary of image morphology, \textit{e.g.,} trends, spikes, and fluctuations, serving as high-level \emph{vision priors}~\citep{liu2025clipchart}. We therefore ask, \emph{can such vision priors reshape and augment MedTS representations learned from raw numerical signals?}

In this work, we answer this question by proposing \textbf{ViRe} (\textbf{V}ision-\textbf{I}nformed \textbf{Re}trieval), a clinically grounded framework that bridges how models and clinicians ``see'' MedTS. ViRe adopts a dual view: \emph{(i) a numerical view} modeled by time series Transformers, and \emph{(ii) a waveform view} encoded by a pre-trained \textbf{\textit{frozen}} CLIP vision encoder. Inspired by vision-to-text alignment in image--text models, we treat the waveform embedding as a \emph{Vision Query} that retrieves \emph{evidence} from the numerical sequence~\citep{jia2021retrieval}. Via a tailored attention-based retrieval, this Vision Query \emph{selects} and aggregates morphology-relevant numerical tokens, while the temporal and channel features remain the retrieved evidence. Across six MedTS benchmarks, ViRe consistently surpasses strong baselines such as Medformer~\citep{wang2024medformer} by \textbf{6.42\%} on average. Our contributions are threefold.
\begin{itemize}[leftmargin=1.5em, topsep=2pt, itemsep=2pt, parsep=0pt, partopsep=0pt]
\item We identify the mismatch between numerics-oriented representation learning of existing MedTS models and waveform-based clinical reasoning, which motivates a morphology-aware prior.
\item We deterministically visualize MedTS as waveforms and extract high-level morphology-aware priors using a pre-trained \textit{frozen} CLIP Vision Encoder without any fine-tuning.
\item ViRe introduces asymmetric dual-path cross-attention, where a \emph{VLM-derived global Vision Query} retrieves temporal and channel evidence from numerical MedTS tokens.
\end{itemize}

\section{Related Work}

\paragraph{Medical Time Series Analysis.}
Recent MedTS classification has progressed from handcrafted feature pipelines to deep representation learning~\citep{fahimi2017index0,Ganapathy2018medts}. Convolutional and recurrent models learn directly from raw EEG/ECG, while recent Transformers further capture long-range temporal dependencies and inter-channel interactions with attention~\citep{arif2024ef,wang2024medformer}.

Yet prevailing methods remain mainly numerical, treating MedTS as sequences or matrices and learning from signal values alone, without any explicit notion of waveform shape. Clinical interpretation is instead morphology-centric: EEG and ECG are reviewed as waveforms, where diagnostic cues are recognized visually and summarized into structured reports and quantitative indices~\citep{libenson2010eegvision,hirsch2021eegvision,kligfield2007ecgvision,nejedly2019eegvision}. This mismatch limits current classifiers; ViRe instead uses visual morphology as a retrieval prior that selects clinically relevant evidence from the numerical representation.

\paragraph{Vision-Language Models for Time Series.}
Vision-Language Models (VLMs) (\textit{e.g.}, CLIP~\citep{radford2021clip}, ViLT~\citep{kim2021vilt}) learn concept-aligned visual embeddings through large-scale image-text pretraining. This makes VLMs attractive for extracting compact morphology-aware priors from waveform plots, whose visual structures encode clinically meaningful patterns. Recent progress on VLMs for time series largely follows three directions. First, some works visualize time series and use vision/VLM representations for forecasting. ViTime~\citep{yang2025vitime} makes the image representation the primary modeling space, while Time-VLM~\citep{zhong2025timevlm} fuses visual/textual augmentations with temporal features. Another line explicitly aligns time series with their textual information in a CLIP-style manner, in which the textual source is processed using CLIP Language Encoder, for zero-shot recognition (\textit{e.g.}, TS-CLIP)~\citep{chen2025tsclip}. Finally, studies also investigate whether VLMs can classify time series competitively when fed with visualizations~\citep{prithyani2024feasibility}. These directions assign distinct roles to visual and textual information: ViTime predicts in an image-centered space, Time-VLM fuses visual/textual augmentations with temporal features, TS-CLIP aligns time series with text, and visualization-based classifiers operate directly on rendered signals. ViRe uses a frozen CLIP waveform representation as a global Vision Query that retrieves temporal and channel tokens, linking visual morphology to numerical evidence.

\section{Methodology}

This section first introduces the preliminaries of the medical time series (MedTS) (\textit{e.g.}, EEG and ECG) classification-related foundations. Then, we present the proposed \textbf{ViRe} (\textbf{V}ision-\textbf{I}nformed \textbf{Re}trieval) framework, which aligns the numerical MedTS signals with their vision priors.

\subsection{Preliminaries}
\define{Problem Formulation} \textit{Considering a MedTS sample $X \in \mathbb{R}^{T\times C}$, where $T$ denotes the number of timestamps and $C$ is the number of channels, our objective is to learn a function (model) that predicts the corresponding label $\widehat{Y} \in \mathbb{R}^{K}$. $K$ denotes the number of classes, e.g., various disease types or different stages of the same disease, depending on the specific diagnostic task.}

MedTS naturally forms a hierarchical structure: each dataset contains multiple subjects (\textit{patients}), whose records are organized into sessions (\textit{clinical visits}), further divided into trials (\textit{repeated measurements}), and finally into short segments (\textit{samples}) that are fed to diagnostic models~\citep{wang2024subin0,tzimourta2021mlmed,Jin2024ecg1,gow2023mimic,wagner2020ptb-xl}. Clinicians, however, make decisions at the \emph{\textbf{subject}} level. To align with the real-world practice, we follow the \textbf{Subject-Independent} setting~\citep{wang2024medformer,miltiadous2023dicenet,escudero2006apava}: \emph{the splitting is performed by subject, and samples from the same patient are assigned exclusively to the training, validation, or test set}. This closely simulates real-world practice, where models must generalize to unseen patients, and therefore provides a clinically meaningful comparison across all methods.

\subsection{The ViRe Framework}
\begin{figure}[!t]
  \centering
  \includegraphics[width=\ViReMethodFigureWidth]{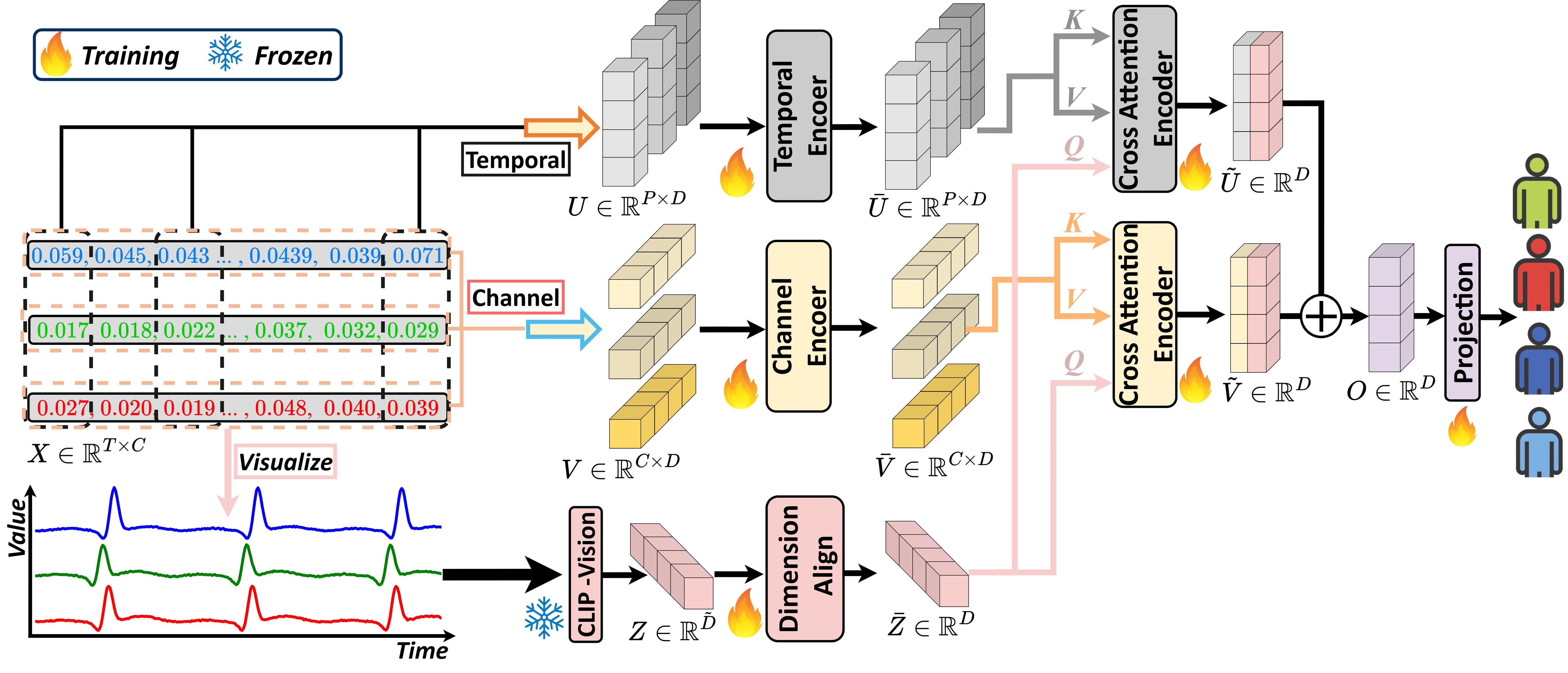}
  \vspace{-2mm}
  \caption{\textbf{Overview of ViRe.} Raw MedTS is embedded into \textit{Temporal} and \textit{Channel} tokens, while its waveform visualization is processed by a \textbf{\textit{frozen}} CLIP Vision Encoder to produce a global \textit{Vision Query}. Cross-attention Encoders retrieve vision-guided temporal and channel summaries, which are fused and projected to produce the final diagnostic prediction.}
  \label{fig:vire}
\end{figure}

The proposed \textbf{ViRe} is illustrated in \figref{fig:vire}. Given a MedTS sample $X \in \mathbb{R}^{T \times C}$, we first tokenize it into the \textbf{Temporal embedding}~\citep{wang2024medformer} and the \textbf{Channel embedding}~\citep{liu2023itransformer} along two complementary axes, \textit{i.e.}, the temporal and channel dimension. Two Transformer Encoders are applied to extract temporal and channel features from these embeddings, respectively. In parallel, the same MedTS is deterministically visualized into a waveform image and fed into a \emph{frozen} \textbf{CLIP Vision Encoder}~\citep{radford2021clip}. The frozen encoder summarizes the rendered waveform into a compact morphology-aware prior for numerical evidence retrieval. ViRe then performs \emph{\textbf{Vision-Informed Retrieval}} to align the vision and numerical modalities: the vision feature is used as the shared \emph{Query} $(Q)$, while the Temporal and Channel features provide \emph{Keys} and \emph{Values} $(K,V)$ to the corresponding cross-attention Encoder. This yields two vision-guided summaries of temporal and channel evidence. Their sum is projected to the classification logits, so the visual representation guides retrieval from both numerical token streams.

The frozen visual branch supplies only the retrieval Query while all Keys and Values remain numerical, and the same Query conditions both tokenizations. Thus, vision selects \emph{what evidence to retrieve} without becoming an independent diagnostic pathway of its own.

\paragraph{Numerical modeling.} ViRe represents each sample $X\in\mathbb{R}^{T\times C}$ through two complementary views: a \textbf{Temporal embedding} that captures dynamics across timestamps and a \textbf{Channel embedding} that preserves channel-wise semantics and explicitly models inter-channel dependencies.

Along the temporal axis, we split the signal into non-overlapping segments (with length $L$) and treat each segment across all channels as a token:
\begin{equation}
\begin{aligned}
U_{i,:} &= \operatorname{vec}\!\big(X_{(i-1)L:iL,:}\big)W_t+b_t+W^{tpos}_{i,:},\\[-1mm]
i&=1,\ldots,P,\quad P=\left\lceil\tfrac{T}{L}\right\rceil,\\[-1mm]
W_t&\in\mathbb{R}^{LC\times D},\quad b_t\in\mathbb{R}^{D},\quad W^{tpos}\in\mathbb{R}^{P\times D}.
\end{aligned}
\end{equation}
where $\operatorname{vec}:\mathbb{R}^{m\times n}\!\to\mathbb{R}^{mn}$ flattens a matrix into a vector. $W^{tpos}$ is the position embedding~\citep{vaswani2017attention}. Stacking all temporal tokens yields $U\in\mathbb{R}^{P\times D}$. Such tokenization eases the modeling of trend and seasonality patterns~\citep{zhang2022tdformer}. Orthogonally, we summarize each channel by aggregating its entire trajectory into a token:
\begin{equation}
\begin{aligned}
V_{j,:}&=X_{:,j}^{\top}W_c+b_c+W^{cpos}_{j,:},\quad j=1,\ldots,C,\\[-1mm]
W_c&\in\mathbb{R}^{T\times D},\quad b_c\in\mathbb{R}^{D},\quad W^{cpos}\in\mathbb{R}^{C\times D}.
\end{aligned}
\end{equation}
With another position embedding $W^{cpos}$, this forms the Channel embedding $V\in\mathbb{R}^{C\times D}$. Summarizing each channel's full trajectory in one token preserves channel-specific semantics, which is crucial for modeling inter-channel correlations~\citep{yue2025freeformer,qiu2025duet,hu2025timefilter}. Separate Transformer Encoders then produce the temporal feature \mbox{$\bar U\in\mathbb{R}^{P\times D}$} and channel feature \mbox{$\bar V\in\mathbb{R}^{C\times D}$} used as Keys and Values in retrieval.

\paragraph{Vision Query.} To inject vision priors, we convert the same $X\in\mathbb{R}^{T\times C}$ into a waveform image via a deterministic visualization operator (detailed in \appref{app:visualization_operator}). It plots and stacks signals from all channels into a single image:
\begin{equation}
I=\mathcal{V}(X)\in\mathbb{R}^{H\times W\times3}.
\end{equation}
The image $I$ is fed into a pre-trained \textit{frozen} \textbf{CLIP Vision Encoder} $f_{\mathrm{CLIP}}(\cdot)$~\citep{radford2021clip}:
\begin{equation}
Z=f_{\mathrm{CLIP}}(I)\in\mathbb{R}^{\tilde D},
\end{equation}
yielding a high-level visual feature that embeds text--image alignment knowledge. A \textbf{\textit{Dimension Align layer}} projects $Z$ into the shared numerical latent space:
\begin{equation}
\bar Z=ZW_z+b_z,\qquad W_z\in\mathbb{R}^{\tilde D\times D},\quad b_z\in\mathbb{R}^{D}.
\end{equation}
\eqparbreak

The aligned feature $\bar Z\in\mathbb{R}^{D}$ serves as the global \textit{Vision Query}. Sharing the latent dimension with $\bar U$ and $\bar V$ lets the query attend to both numerical views through a common cross-attention layer.

\paragraph{Cross-Modal Alignment.} Given the Temporal embedding $\bar U\in\mathbb{R}^{P\times D}$, Channel embedding $\bar V\in\mathbb{R}^{C\times D}$, and the vision-derived Query $\bar Z\in\mathbb{R}^{D}$, ViRe performs Cross-Modal Alignment via two cross-attention Encoders~\citep{liu2025timebridge,wang2024timexer,liu2024timecma}. We reshape the Vision Query into a shared token $Q\in\mathbb{R}^{1\times D}$; the Temporal and Channel embeddings provide the following two numerical Key--Value sets:
\begin{equation}
\begin{aligned}
\widetilde U &= \mathrm{CrossAttn}(Q,\bar U,\bar U)\in\mathbb{R}^{1\times D},\\[-1mm]
\widetilde V &= \mathrm{CrossAttn}(Q,\bar V,\bar V)\in\mathbb{R}^{1\times D}.
\end{aligned}
\end{equation}
We add the two retrieved summaries and project the result:
\begin{equation}
\begin{aligned}
O &= \widetilde U+\widetilde V\in\mathbb{R}^{1\times D},\\[-1mm]
\widehat Y &= OW_y+b_y,\qquad W_y\in\mathbb{R}^{D\times K},\ b_y\in\mathbb{R}^{K}.
\end{aligned}
\end{equation}
producing the final logits $\widehat Y\in\mathbb{R}^{K}$ for MedTS diagnosis. The same query is shared by both retrieval paths, so their outputs remain directly comparable before fusion while each path attends over a different numerical organization of the same underlying input waveform.

Since we employ a pre-trained \textit{frozen} CLIP Vision Encoder as the vision feature extractor, training requires no auxiliary objectives. We use the standard classification objective, namely cross-entropy between the predicted labels $\widehat Y$ and the ground-truth $Y$~\citep{afzal2024rest,chen2022me}. During training, one augmentation operation per input is applied consistently to the signal and its rendering (\appref{app:augmentation}); at validation and test time, only the deterministic rendering pipeline is used.

\section{Experimental Results}
\subsection{Experiment Setting}
\subsubsection{Datasets}
We evaluate six subject-disjoint public benchmarks spanning EEG and ECG. \textbf{APAVA}~\citep{escudero2006apava} and \textbf{ADFTD}~\citep{miltiadous2023adftd} cover Alzheimer's-related EEG classification; \textbf{TDBrain}~\citep{van2022tdbrain} targets Parkinson's disease from EEG. For ECG, \textbf{PTB}~\citep{physiobank2000ptb} evaluates myocardial infarction, \textbf{PTB-XL}~\citep{wagner2020ptb-xl} provides five diagnostic classes, and \textbf{MIMIC}~\citep{gow2023mimic} evaluates heart disease versus healthy controls.

Splits are subject-disjoint, so every test sample comes from an unseen individual. \tabref{tab:data_inf} summarizes the statistics, while \appref{app:data} gives segmentation and split details. The six benchmarks span EEG and ECG, varied cohort sizes, and different channel configurations, testing one retrieval mechanism without dataset-specific changes. EEG recordings are segmented into 1-second windows at 256 Hz, PTB-XL into 1-second windows at 250 Hz, and PTB and MIMIC into R-peak-aligned single heartbeats, following the Medformer preprocessing protocol~\citep{wang2024medformer}.
\begingroup
\hbadness=10001
\par\medskip
\noindent\begin{minipage}{\textwidth}
\captionsetup{type=table,font=small,skip=2pt,justification=centering,singlelinecheck=false}
\captionof{table}{\textbf{Dataset statistics after preprocessing.} Length is the input sequence length.}
\label{tab:data_inf}
\renewcommand{\arraystretch}{1.02}
\setlength{\tabcolsep}{0pt}
\small
\begin{tabular*}{\textwidth}{@{\extracolsep{\fill}} l l r r r r r @{} }
\toprule
\textbf{Modality} & \textbf{Dataset} & \textbf{Subjects} & \textbf{Samples} & \textbf{Classes} & \textbf{Channels} & \textbf{Length} \\
\midrule
\multirow{3}{*}{EEG} & APAVA   & 23     & 5,967   & 2 & 16 & 256 \\
                      & ADFTD   & 88     & 69,752  & 3 & 19 & 256 \\
                      & TDBrain & 72     & 6,240   & 2 & 33 & 256 \\
\addlinespace[2pt]
\multirow{3}{*}{ECG} & PTB     & 198    & 64,356  & 2 & 15 & 300 \\
                      & PTB-XL  & 17,596 & 191,400 & 5 & 12 & 250 \\
                      & MIMIC   & 20,437 & 204,370 & 2 & 12 & 250 \\
\bottomrule
\end{tabular*}
\end{minipage}\par
\endgroup

\subsubsection{Baselines}
We compare ViRe with ten representative architectures spanning complementary design families: Autoformer and FEDformer for decomposition/frequency modeling~\citep{wu2021autoformer,zhou2022fedformer}; Informer, Reformer, and Transformer for attention variants~\citep{zhou2021informer,kitaev2019reformer,vaswani2017attention}; MTST, Nonformer, iTransformer, and PatchTST for patch-, non-stationary-, and channel-aware modeling~\citep{zhang2024mtst,liu2022nonfromer,liu2023itransformer,nie2022patchtst}; and Medformer as the dedicated state-of-the-art baseline for medical time series classification~\citep{wang2024medformer}. All baselines share the same encoder depth, model dimension, and optimizer (\appref{app:baseline_implementation}).

\noindent\begin{minipage}{\textwidth}
\begin{center}
\captionsetup{type=table,font=small,skip=5pt}
\captionof{table}{\textbf{Subject-independent classification summary.} Per-dataset Avg over Accuracy, Precision, Recall, F1, AUROC, and AUPRC. Full mean$\pm$std results are in \appref{app:full_results}; \first{best} is bolded and \second{second} is blue italics.}
\vspace{-1.6mm}
\label{tab:bench_summary}
\renewcommand{\arraystretch}{1.03}
\setlength{\tabcolsep}{5.2pt}
\resizebox{0.85\textwidth}{!}{%
\begin{tabular}{lrrrrrrr}
\toprule
& \multicolumn{3}{c}{\textbf{EEG}} & \multicolumn{3}{c}{\textbf{ECG}} & \\
\cmidrule(lr){2-4}\cmidrule(lr){5-7}
\textbf{Model} & \textbf{APAVA} & \textbf{ADFTD} & \textbf{TDBrain} & \textbf{PTB} & \textbf{PTB-XL} & \textbf{MIMIC} & \textbf{Mean} \\
\midrule
Autoformer   & 70.71 & 46.43 & 89.52 & 70.86 & 57.53 & 79.69 & 69.12 \\
FEDformer    & 77.21 & 48.20 & 80.98 & 75.90 & 56.83 & 86.79 & 70.98 \\
Informer     & 71.36 & 50.04 & 91.64 & 80.62 & 67.84 & 86.94 & 74.74 \\
iTransformer & 77.23 & 51.71 & 77.63 & \second{84.95} & 64.45 & 87.11 & 73.85 \\
MTST         & 69.94 & 47.89 & 79.35 & 76.80 & 68.70 & 87.52 & 71.70 \\
Nonformer    & 70.70 & 50.94 & 91.07 & 79.58 & 66.78 & 86.35 & 74.24 \\
PatchTST     & 65.89 & 45.56 & 81.94 & 75.35 & \first{69.90} & 86.79 & 70.91 \\
Reformer     & 77.02 & 53.19 & 90.84 & 79.51 & 68.04 & \second{87.90} & 76.08 \\
Transformer  & 74.42 & 52.09 & 90.34 & 78.69 & 66.73 & 87.01 & 74.88 \\
Medformer    & \second{79.74} & \second{54.63} & \second{91.91} & 84.69 & 69.28 & 87.14 & \second{77.90} \\
\addlinespace[1pt]
\rowcolor{softband}\textbf{ViRe} & \first{92.79} & \first{59.83} & \first{95.54} & \first{89.08} & \second{69.47} & \first{90.68} & \first{82.90} \\
\bottomrule
\end{tabular}%
}
\end{center}
\end{minipage}\par

\subsubsection{Implementation}
We report Accuracy, Precision, Recall, F1-Score, AUROC, and AUPRC, use F1-Score for early stopping, and average five random seeds. All experiments run on one NVIDIA RTX 4090 GPU. Baselines are reproduced through the Medformer benchmark~\citep{wang2024medformer} using identical subject-level splits, metrics, and early-stopping criteria. Model selection uses validation subjects only; full hyperparameters, preprocessing details, and code availability are given in \appref{app:implementation}.

\subsection{Classification Performance}
\tabref{tab:bench_summary} provides one Avg per dataset, defined as the arithmetic mean of the six classification metrics. ViRe ranks first on five of six benchmarks. Relative to Medformer, Avg improves from 79.74 to 92.79 on APAVA (\textbf{16.37\%}), 54.63 to 59.83 on ADFTD (\textbf{9.52\%}), 91.91 to 95.54 on TDBrain (\textbf{3.95\%}), 84.69 to 89.08 on PTB (\textbf{5.18\%}), and 87.14 to 90.68 on MIMIC (\textbf{4.06\%}). On PTB-XL, PatchTST leads with 69.90, followed closely by ViRe at 69.47 and Medformer at 69.28.

Across all six datasets, ViRe reaches an overall mean Avg of \textbf{82.90}, versus 77.90 for Medformer and 76.08 for Reformer. Against the strongest non-ViRe entry, ViRe leads APAVA, ADFTD, TDBrain, PTB, and MIMIC by 13.05, 5.20, 3.63, 4.13, and 2.78 points, respectively; on PTB-XL it ranks second, only 0.43 points below PatchTST. Full metric-wise mean$\pm$std results are provided in \appref{app:full_results}.

The gains span both EEG and ECG and persist from small cohorts to larger benchmarks, indicating that the benefit is not tied to a single modality or dataset scale. PTB-XL is the main exception: ViRe remains competitive but does not surpass PatchTST, suggesting that morphology-aware retrieval complements rather than uniformly dominates strong numerical patch modeling.

Two cross-dataset patterns are especially informative. The largest gains occur on APAVA and ADFTD, the two smaller EEG cohorts, where a stable morphology prior can be valuable when subject-level supervision is limited. At the same time, positive improvements persist on TDBrain, PTB, and MIMIC despite substantial differences in channel count, cohort size, and signal modality. Together with the near-tie on PTB-XL, this pattern supports ViRe as an inductive bias for evidence selection rather than a dataset-specific capacity increase: its benefit is strongest when waveform morphology helps isolate diagnostic evidence that a numerical encoder alone may not emphasize.

\subsection{Model Analysis}
We isolate where the gain comes from by varying one factor at a time while keeping the numerical pipeline fixed: \textbf{(i)} the retrieval Query, which tests whether the benefit stems from the retrieval mechanism or from the VLM-derived Query content; \textbf{(ii)} the fusion operator, comparing cross-attention retrieval with addition and concatenation; \textbf{(iii)} the vision backbone, which contrasts random, ImageNet, and CLIP initializations; and \textbf{(iv)} the training-data proportion, which probes data efficiency. Studies (i)--(iii) use two EEG (ADFTD, APAVA) and two ECG (PTB, MIMIC) datasets and report Accuracy, F1-Score, and the average relative improvement over the variant without retrieval (w/o).
\subsubsection{Ablation of Retrieval Query}
\begin{table}[H]
\centering
\def\arraystretch{1.0}
\caption{\textbf{\textit{Ablation of the retrieval Query.}} Zero replaces the Vision Query with an all-zero vector, and Gaussian replaces it with a vector sampled from a Gaussian distribution.}
\vspace{-3mm}
\label{tab:ab_query}
\resizebox{0.96\textwidth}{!}{
\begin{tabular}{cccccccccc}

    \toprule
    & \multicolumn{2}{c}{\scalebox{1.3}{\textbf{ADFTD}}} & \multicolumn{2}{c}{\scalebox{1.3}{\textbf{APAVA}}} & \multicolumn{2}{c}{\scalebox{1.3}{\textbf{PTB}}} & \multicolumn{2}{c}{\scalebox{1.3}{\textbf{MIMIC}}} &  \\
    \cmidrule[1pt](l{10pt}r{10pt}){2-3} \cmidrule[1pt](l{10pt}r{10pt}){4-5}\cmidrule[1pt](l{10pt}r{10pt}){6-7}\cmidrule[1pt](l{10pt}r{10pt}){8-9}
    
    \scalebox{1.3}{\textbf{Query Type}} & \multicolumn{1}{c}{\scalebox{1.3}{\textbf{Accuracy}}} & \multicolumn{1}{c}{\scalebox{1.3}{\textbf{F1-Score}}} & \multicolumn{1}{c}{\scalebox{1.3}{\textbf{Accuracy}}} & \multicolumn{1}{c}{\scalebox{1.3}{\textbf{F1-Score}}} & \multicolumn{1}{c}{\scalebox{1.3}{\textbf{Accuracy}}} & \multicolumn{1}{c}{\scalebox{1.3}{\textbf{F1-Score}}}& \multicolumn{1}{c}{\scalebox{1.3}{\textbf{Accuracy}}} & \multicolumn{1}{c}{\scalebox{1.3}{\textbf{F1-Score}}} & \scalebox{1.3}{\textbf{Avg. Gain}}\\
    \toprule[1pt]

    \scalebox{1.3}{\textbf{w/o}}
    & \scalebox{1.3}{54.79\std{1.81}} & \scalebox{1.3}{51.73\std{2.22}} & \scalebox{1.3}{83.86\std{2.87}} & \scalebox{1.3}{82.45\std{3.69}} & \scalebox{1.3}{81.45\std{3.44}} & \scalebox{1.3}{74.58\std{6.89}} & \scalebox{1.3}{84.92\std{1.11}} & \scalebox{1.3}{84.81\std{1.12}} & \scalebox{1.3}{\text{--}} \\

    \scalebox{1.3}{\textbf{Zero}}
    & \scalebox{1.3}{54.72\std{2.82}} & \scalebox{1.3}{51.23\std{2.72}} & \scalebox{1.3}{84.34\std{1.72}} & \scalebox{1.3}{83.73\std{1.96}} & \scalebox{1.3}{84.62\std{2.55}} & \scalebox{1.3}{80.21\std{3.96}} & \scalebox{1.3}{86.13\std{0.14}} & \scalebox{1.3}{86.06\std{0.13}} & \scalebox{1.3}{1.92\%} \\

    \scalebox{1.3}{\textbf{Gaussian}}
    & \scalebox{1.3}{54.46\std{2.21}} & \scalebox{1.3}{51.54\std{1.93}} & \scalebox{1.3}{82.89\std{2.66}} & \scalebox{1.3}{81.65\std{3.12}} & \scalebox{1.3}{82.83\std{2.15}} & \scalebox{1.3}{77.34\std{3.64}} & \scalebox{1.3}{86.50\std{0.11}} & \scalebox{1.3}{86.44\std{0.11}} & \scalebox{1.3}{0.76\%} \\

    \scalebox{1.3}{\textbf{Vision}}
    & \scalebox{1.3}{\first{57.83\std{1.82}}} & \scalebox{1.3}{\first{54.02\std{2.22}}} & \scalebox{1.3}{\first{91.43\std{1.03}}} & \scalebox{1.3}{\first{91.19\std{1.03}}} & \scalebox{1.3}{\first{88.26\std{1.12}}} & \scalebox{1.3}{\first{85.81\std{1.52}}} & \scalebox{1.3}{\first{88.61\std{0.12}}} & \scalebox{1.3}{\first{88.54\std{0.12}}} & \scalebox{1.3}{\first{7.72\%}} \\

\bottomrule[1pt]
\end{tabular}
}
\vspace{-3mm}
\end{table}

\tabref{tab:ab_query} evaluates the impact of different retrieval Query. Removing retrieval (w/o) gives the worst results. Content-free retrieval queries provide modest gains: an all-zero Query (Zero) improves the average by 1.92\%, and a Gaussian-initialized Query (Gaussian) by 0.76\%. The Vision-Informed Query, derived from the VLM vision-language space, achieves the strongest improvement (7.72\%). For example, on APAVA, Accuracy increases from 83.86 to 91.43 (9.03\%) and F1 from 82.45 to 91.19 (10.60\%). \textbf{\textit{The results identify the Vision Query as the main source of the retrieval gain, providing a structured VLM-derived prior for evidence selection.}}

\subsubsection{Ablation of Modality Fusion}
\begin{table}[H]
\centering
\def\arraystretch{1.0}
\caption{\textbf{\textit{Ablation of modality fusion.}} Cross-attention retrieval is compared with simple addition (\textit{Add}) and concatenation (\textit{Concat}) of the vision prior and the numerical features.}
\vspace{-3mm}
\label{tab:ab_fuse}
\resizebox{1\textwidth}{!}{
\begin{tabular}{cccccccccc}

    \toprule
    & \multicolumn{2}{c}{\scalebox{1.3}{\textbf{ADFTD}}} & \multicolumn{2}{c}{\scalebox{1.3}{\textbf{APAVA}}} & \multicolumn{2}{c}{\scalebox{1.3}{\textbf{PTB}}} & \multicolumn{2}{c}{\scalebox{1.3}{\textbf{MIMIC}}} &  \\
    \cmidrule[1pt](l{10pt}r{10pt}){2-3} \cmidrule[1pt](l{10pt}r{10pt}){4-5}\cmidrule[1pt](l{10pt}r{10pt}){6-7}\cmidrule[1pt](l{10pt}r{10pt}){8-9}
    
    \scalebox{1.3}{\textbf{Fusion Strategy}} & \multicolumn{1}{c}{\scalebox{1.3}{\textbf{Accuracy}}} & \multicolumn{1}{c}{\scalebox{1.3}{\textbf{F1-Score}}} & \multicolumn{1}{c}{\scalebox{1.3}{\textbf{Accuracy}}} & \multicolumn{1}{c}{\scalebox{1.3}{\textbf{F1-Score}}} & \multicolumn{1}{c}{\scalebox{1.3}{\textbf{Accuracy}}} & \multicolumn{1}{c}{\scalebox{1.3}{\textbf{F1-Score}}}& \multicolumn{1}{c}{\scalebox{1.3}{\textbf{Accuracy}}} & \multicolumn{1}{c}{\scalebox{1.3}{\textbf{F1-Score}}} & \scalebox{1.3}{\textbf{Avg. Gain}}\\
    \toprule[1pt]

    \scalebox{1.3}{\textbf{w/o}}
    & \scalebox{1.3}{54.79\std{1.81}} & \scalebox{1.3}{51.73\std{2.22}} & \scalebox{1.3}{83.86\std{2.87}} & \scalebox{1.3}{82.45\std{3.69}} & \scalebox{1.3}{81.45\std{3.44}} & \scalebox{1.3}{74.58\std{6.89}} & \scalebox{1.3}{84.92\std{1.11}} & \scalebox{1.3}{84.81\std{1.12}} & \scalebox{1.3}{\text{--}} \\

    \scalebox{1.3}{\textbf{Add}}
    & \scalebox{1.3}{56.67\std{1.46}} & \scalebox{1.3}{53.53\std{2.12}} & \scalebox{1.3}{85.95\std{1.77}} & \scalebox{1.3}{85.45\std{1.71}} & \scalebox{1.3}{84.81\std{2.17}} & \scalebox{1.3}{81.26\std{3.47}} & \scalebox{1.3}{87.57\std{0.06}} & \scalebox{1.3}{87.49\std{0.07}} & \scalebox{1.3}{4.05\%} \\

    \scalebox{1.3}{\textbf{Concat}}
    & \scalebox{1.3}{56.86\std{1.54}} & \scalebox{1.3}{53.55\std{1.21}} & \scalebox{1.3}{84.23\std{2.55}} & \scalebox{1.3}{83.02\std{3.11}} & \scalebox{1.3}{83.60\std{2.25}} & \scalebox{1.3}{79.13\std{3.47}} & \scalebox{1.3}{87.88\std{0.18}} & \scalebox{1.3}{87.81\std{0.18}} & \scalebox{1.3}{3.02\%} \\

    \scalebox{1.3}{\textbf{Retrieval}}
    & \scalebox{1.3}{\first{57.83\std{1.82}}} & \scalebox{1.3}{\first{54.02\std{2.22}}} & \scalebox{1.3}{\first{91.43\std{1.03}}} & \scalebox{1.3}{\first{91.19\std{1.03}}} & \scalebox{1.3}{\first{88.26\std{1.12}}} & \scalebox{1.3}{\first{85.81\std{1.52}}} & \scalebox{1.3}{\first{88.61\std{0.12}}} & \scalebox{1.3}{\first{88.54\std{0.12}}} & \scalebox{1.3}{\first{7.72\%}} \\

\bottomrule[1pt]
\end{tabular}
}
\vspace{-3mm}
\end{table}

\tabref{tab:ab_fuse} studies how different fusion strategies impact ViRe. The w/o variant simply adds the Temporal and Channel branches without using external priors, and gives the poorest performance. Replacing it with naive multimodal fusion already helps: both simple Add (4.05\%) and Concat (3.02\%) consistently outperform w/o on all datasets. Add and Concat show that vision priors already improve numerical modeling under simple fusion. Cross-attention retrieval further raises the overall gain to 7.72\%. Its token-wise, content-dependent interaction provides the strongest integration of vision priors with numerical temporal and channel evidence across all four datasets.

\subsubsection{Ablation of Vision Backbone}
\begin{table}[H]
\centering
\def\arraystretch{1.0}
\caption{\textbf{\textit{Ablation of the vision backbone:}} Random ViT, ImageNet ViT, and CLIP-Vision.}
\vspace{-3mm}
\label{tab:ab_vit}
\resizebox{1\textwidth}{!}{
\begin{tabular}{cccccccccc}

    \toprule
    & \multicolumn{2}{c}{\scalebox{1.3}{\textbf{ADFTD}}} & \multicolumn{2}{c}{\scalebox{1.3}{\textbf{APAVA}}} & \multicolumn{2}{c}{\scalebox{1.3}{\textbf{PTB}}} & \multicolumn{2}{c}{\scalebox{1.3}{\textbf{MIMIC}}} &  \\
    \cmidrule[1pt](l{10pt}r{10pt}){2-3} \cmidrule[1pt](l{10pt}r{10pt}){4-5}\cmidrule[1pt](l{10pt}r{10pt}){6-7}\cmidrule[1pt](l{10pt}r{10pt}){8-9}
    
    \scalebox{1.3}{\textbf{Vision Backbone}} & \multicolumn{1}{c}{\scalebox{1.3}{\textbf{Accuracy}}} & \multicolumn{1}{c}{\scalebox{1.3}{\textbf{F1-Score}}} & \multicolumn{1}{c}{\scalebox{1.3}{\textbf{Accuracy}}} & \multicolumn{1}{c}{\scalebox{1.3}{\textbf{F1-Score}}} & \multicolumn{1}{c}{\scalebox{1.3}{\textbf{Accuracy}}} & \multicolumn{1}{c}{\scalebox{1.3}{\textbf{F1-Score}}}& \multicolumn{1}{c}{\scalebox{1.3}{\textbf{Accuracy}}} & \multicolumn{1}{c}{\scalebox{1.3}{\textbf{F1-Score}}} & \scalebox{1.3}{\textbf{Avg. Gain}}\\
    \toprule[1pt]

    \scalebox{1.3}{\textbf{w/o}}
    & \scalebox{1.3}{54.79\std{1.81}} & \scalebox{1.3}{51.73\std{2.22}} & \scalebox{1.3}{83.86\std{2.87}} & \scalebox{1.3}{82.45\std{3.69}} & \scalebox{1.3}{81.45\std{3.44}} & \scalebox{1.3}{74.58\std{6.89}} & \scalebox{1.3}{84.92\std{1.11}} & \scalebox{1.3}{84.81\std{1.12}} & \scalebox{1.3}{\text{--}} \\
    
    \scalebox{1.3}{\textbf{ViT (Random Init)}}
    & \scalebox{1.3}{37.39\std{3.11}} & \scalebox{1.3}{27.34\std{1.92}}& \scalebox{1.3}{80.62\std{0.60}}& \scalebox{1.3}{77.18\std{0.54}}& \scalebox{1.3}{80.65\std{1.68}}& \scalebox{1.3}{73.50\std{2.79}}& \scalebox{1.3}{83.67\std{0.14}} & \scalebox{1.3}{83.58\std{0.13}} & \scalebox{1.3}{-11.81\%} \\

    \scalebox{1.3}{\textbf{ViT (ImageNet)}}
    & \scalebox{1.3}{52.21\std{0.73}} & \scalebox{1.3}{50.55\std{1.13}}& \scalebox{1.3}{84.48\std{2.80}}& \scalebox{1.3}{84.86\std{3.67}}& \scalebox{1.3}{82.06\std{3.54}}& \scalebox{1.3}{75.72\std{5.83}}& \scalebox{1.3}{86.52\std{0.21}}& \scalebox{1.3}{86.42\std{0.21}} & \scalebox{1.3}{0.34\%} \\

    \scalebox{1.3}{\textbf{CLIP-Vision}}
    & \scalebox{1.3}{\first{57.83\std{1.82}}} & \scalebox{1.3}{\first{54.02\std{2.22}}} & \scalebox{1.3}{\first{91.43\std{1.03}}} & \scalebox{1.3}{\first{91.19\std{1.03}}} & \scalebox{1.3}{\first{88.26\std{1.12}}} & \scalebox{1.3}{\first{85.81\std{1.52}}} & \scalebox{1.3}{\first{88.61\std{0.12}}} & \scalebox{1.3}{\first{88.54\std{0.12}}} & \scalebox{1.3}{\first{7.72\%}} \\

\bottomrule[1pt]
\end{tabular}
}
\vspace{-3mm}
\end{table}

\tabref{tab:ab_vit} ablates how different vision backbones affect ViRe. \textit{\textbf{Simply plugging in a generic vision backbone (\textit{e.g.}, ViT) is insufficient for our task.}} Using a randomly initialized ViT even \emph{hurts} performance, leading to an average degradation of 11.81\%. Pretraining ViT on ImageNet recovers some performance but yields only a marginal gain (0.34\%), indicating that pure-image pretraining brings limited benefit for our sequence-waveform alignment objective. In contrast, employing the \textbf{\textit{CLIP Vision Encoder}}, which is pretrained on text-image pairing tasks, delivers a clear advantage: it achieves the best performance on all datasets, improves the average performance by 7.72\%, and outperforms the ImageNet-pretrained ViT by 7.11\%. These results demonstrate that the gains of ViRe stem neither from the ViT architecture alone nor from generic vision pretraining, but from the \emph{\textbf{cross-modal text-image alignment}} learned by VLMs, which forms a well-aligned vision-language latent space and produces human-describable vision priors for morphology-aware retrieval.

\subsubsection{Generalizability Analysis}
\begin{figure}[H]
\centering
\includegraphics[width=0.7\textwidth]{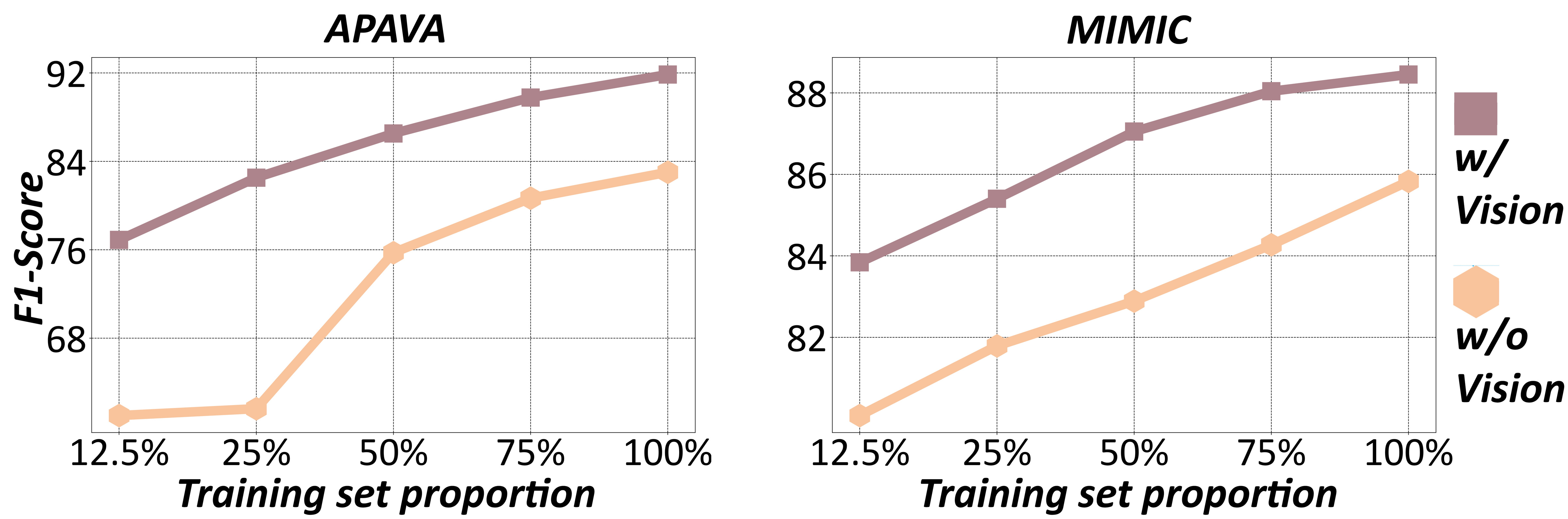}
\captionsetup{skip=4pt}
\caption{F1-Score with and without vision priors across training-set proportions.}
\vspace{-2mm}
\label{fig:ratio}
\end{figure}
We vary the training-set proportion without changing the model to probe the generalizability of the vision priors. \figref{fig:ratio} shows that vision priors consistently improve F1-Score under all data regimes, with the largest separation under limited supervision. The frozen morphology prior therefore improves data efficiency and reduces reliance on abundant labeled subjects.

\subsection{Further Analysis of Vision Feature}
\subsubsection{Analysis of Retrieved Features}
\begin{table}[H]
\small
\centering
\caption{\textbf{Clustering quality of retrieved features.} Raw denotes no retrieval. Lower DBI and higher NMI, Homogeneity, and Completeness indicate better clustering.}\vspace{-3mm}
\resizebox{1\textwidth}{!}{
\begin{threeparttable}
\begin{tabular}{ccccccccccccc}
\toprule
&\multicolumn{4}{c}{\scalebox{1.3}{\textbf{ADFTD}}} & \multicolumn{4}{c}{\scalebox{1.3}{\textbf{PTB}}} & \multicolumn{4}{c}{\scalebox{1.3}{\textbf{MIMIC}}} \\

\cmidrule[1pt](l{10pt}r{10pt}){2-5}\cmidrule[1pt](l{10pt}r{10pt}){6-9}\cmidrule[1pt](l{10pt}r{10pt}){10-13}

\scalebox{1.3}{\textbf{\textit{Metrics/Features}}} 
&\multicolumn{1}{c}{\scalebox{1.3}{\textbf{Vision}}}&\multicolumn{1}{c}{\scalebox{1.3}{\textbf{Zero}}}&\multicolumn{1}{c}{\scalebox{1.3}{\textbf{Gaussian}}}&\multicolumn{1}{c}{\scalebox{1.3}{\textbf{Raw}}}
&\multicolumn{1}{c}{\scalebox{1.3}{\textbf{Vision}}}&\multicolumn{1}{c}{\scalebox{1.3}{\textbf{Zero}}}&\multicolumn{1}{c}{\scalebox{1.3}{\textbf{Gaussian}}}&\multicolumn{1}{c}{\scalebox{1.3}{\textbf{Raw}}}
&\multicolumn{1}{c}{\scalebox{1.3}{\textbf{Vision}}}&\multicolumn{1}{c}{\scalebox{1.3}{\textbf{Zero}}}&\multicolumn{1}{c}{\scalebox{1.3}{\textbf{Gaussian}}}&\multicolumn{1}{c}{\scalebox{1.3}{\textbf{Raw}}}\\ 
\toprule

\scalebox{1.3}{\textbf{DBI \textcolor{red}{$\downarrow$}}} & \scalebox{1.3}{\first{4.034}} & \scalebox{1.3}{7.962} & \scalebox{1.3}{18.098}& \scalebox{1.3}{9.607}
& \scalebox{1.3}{\first{1.156}}& \scalebox{1.3}{1.466}& \scalebox{1.3}{1.722}& \scalebox{1.3}{1.385}
& \scalebox{1.3}{\first{0.893}}& \scalebox{1.3}{1.216}& \scalebox{1.3}{32.475}& \scalebox{1.3}{1.562}\\

\scalebox{1.3}{\textbf{NMI \textcolor{blue}{$\uparrow$}}} & \scalebox{1.3}{\first{0.101}} & \scalebox{1.3}{0.054}& \scalebox{1.3}{0.004}& \scalebox{1.3}{0.042}
& \scalebox{1.3}{\first{0.425}}& \scalebox{1.3}{0.345}& \scalebox{1.3}{0.117}& \scalebox{1.3}{0.194}
& \scalebox{1.3}{\first{0.398}}& \scalebox{1.3}{0.327}& \scalebox{1.3}{0.002}& \scalebox{1.3}{0.276}\\

\scalebox{1.3}{\textbf{Homogeneity \textcolor{blue}{$\uparrow$}}} & \scalebox{1.3}{\first{0.112}} & \scalebox{1.3}{0.055}& \scalebox{1.3}{0.004}& \scalebox{1.3}{0.040}
& \scalebox{1.3}{\first{0.441}}& \scalebox{1.3}{0.371}& \scalebox{1.3}{0.122}& \scalebox{1.3}{0.202}
& \scalebox{1.3}{\first{0.403}}& \scalebox{1.3}{0.332}& \scalebox{1.3}{0.002}& \scalebox{1.3}{0.274}\\

\scalebox{1.3}{\textbf{Completeness \textcolor{blue}{$\uparrow$}}} & \scalebox{1.3}{\first{0.102}} & \scalebox{1.3}{0.053}& \scalebox{1.3}{0.003}& \scalebox{1.3}{0.039}
& \scalebox{1.3}{\first{0.411}}& \scalebox{1.3}{0.332}& \scalebox{1.3}{0.113}& \scalebox{1.3}{0.186}
& \scalebox{1.3}{\first{0.401}}& \scalebox{1.3}{0.329}& \scalebox{1.3}{0.003}& \scalebox{1.3}{0.281}\\

\bottomrule
\end{tabular}
\end{threeparttable}
}
\vspace{-3mm}
\label{tab:cluster}
\end{table}
\begin{figure}[H]
\centering
\includegraphics[width=0.95\textwidth]{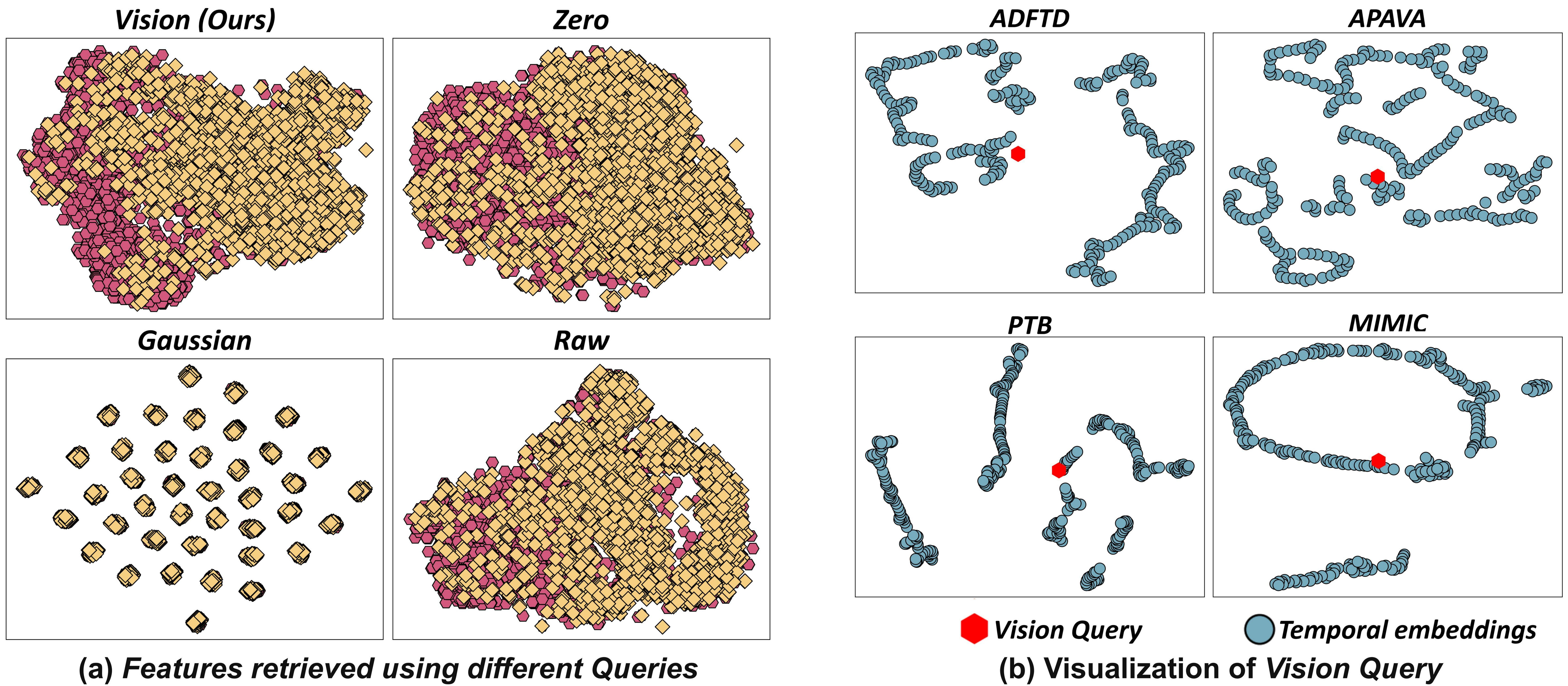}
\captionsetup{skip=4pt}
\caption{\textbf{(a)} t-SNE visualization of retrieved features under different queries. \textbf{(b)} t-SNE geometry of the Vision Query and the temporal embeddings across four datasets.}
\vspace{-2mm}
\label{fig:feature}
\end{figure}
\tabref{tab:cluster} and \figref{fig:feature}\textbf{(a)} analyze the structure of the retrieved feature space under different Query types. Quantitatively, the vision-derived features consistently yield the lowest DBI and the highest NMI, Homogeneity, and Completeness, indicating well-formed and label-consistent clusters. In contrast, the Zero and Gaussian Queries substantially degrade clustering quality, with DBI increasing sharply and the supervised scores dropping close to zero in several cases, especially on ADFTD and MIMIC. The Raw features (no retrieval) preserve some structure but are clearly inferior to the vision-derived features. The t-SNE plots in \figref{fig:feature}\textbf{(a)} echo these trends: with the \textbf{\textit{Vision Query}}, samples from different classes form compact, well-separated manifolds, whereas Raw features exhibit large overlapping regions. The Zero Query only partially improves separability, and the Gaussian Query destroys the class structure, producing a scattered and highly overlapped representation space.

\subsubsection{Geometric Insight into the Vision Query}
\figref{fig:feature}\textbf{(b)} provides a geometric view of how the \textit{Vision Query} interacts with the temporal embeddings. In all cases, the \textit{Vision Query} is not an outlier; instead, it is embedded near the densest region of the representation manifold across datasets. This indicates that the vision-derived Query lies in a semantically meaningful region of the MedTS representation space, making it well-positioned to attend to and aggregate temporal information. Combined with the clustering results in \tabref{tab:cluster}, these findings suggest that the \textit{Vision Query} provides a structured and semantically aligned entry point for cross-attention over numerical embeddings, and that retrieval emphasizes regions that are already informative in the numerical manifold rather than arbitrary directions of the space.

\subsubsection{Attention Alignment with Clinical Morphology}

\begin{table}[H]
\centering
\caption{\textbf{Quantitative Attention Alignment on ECG Datasets.} MAR and CAR measure attention-density enrichment on high-curvature and QRS regions, respectively; higher is better.}\vspace{-3mm}
\label{tab:attention_alignment}
\small
\setlength{\tabcolsep}{11pt}
\renewcommand{\arraystretch}{1.08}
\resizebox{0.70\textwidth}{!}{%
\begin{tabular}{lcccc}
\toprule
& \multicolumn{2}{c}{\textbf{MAR $\uparrow$}} & \multicolumn{2}{c}{\textbf{CAR $\uparrow$}} \\
\cmidrule(lr){2-3}\cmidrule(lr){4-5}
\textbf{Dataset} & \textbf{ViRe} & \textbf{Gaussian} & \textbf{ViRe} & \textbf{Gaussian} \\
\midrule
PTB    & \first{1.81} & 1.46 & \first{1.65} & 1.33 \\
PTB-XL & \first{1.67} & 1.38 & \first{1.51} & 1.27 \\
MIMIC  & \first{1.72} & 1.41 & \first{1.63} & 1.35 \\
\bottomrule
\end{tabular}
}
\vspace{-3mm}
\end{table}

We further quantify whether the retrieval attention concentrates on meaningful temporal regions of ECG. Following \appref{app:attention_alignment}, the Morphology-Aware Attention Ratio (MAR) measures attention-density enrichment on high-curvature timestamps, and the Clinical Alignment Ratio (CAR) measures enrichment on the clinically recognized QRS complex; both divide the attention density inside the target region by that in its complement, so values above one indicate concentration. As reported in \tabref{tab:attention_alignment}, the CLIP-derived Vision Query consistently exceeds a non-semantic Gaussian Query on both metrics across PTB, PTB-XL, and MIMIC. The qualitative map in \appref{app:retrieval_attention} shows the same behavior on individual samples, where attention peaks at synchronous cross-lead variations.

\subsection{Additional Clinical Evidence}
\textbf{PTB-XL diagnosis results and patient-matched controls.} On the official PTB-XL test set, ViRe improves 11 of 13 diagnoses ($p=0.0225$), with the largest gains on HYP and LVH. Zero masking of the visual input reduces macro-F1 by 6.19 points, while cross-patient visual shuffling produces a 15.61-point reduction (\tabref{tab:clinical_ptbxl}). Positive directions also span infarction, ST--T change, ischemia, and rhythm-related categories, extending the pattern beyond HYP and LVH. Repeated subject-disjoint splits confirm this pattern: ViRe wins 7/7 splits on PTB-XL, 8/10 on APAVA, and 9/10 on PTB, and the paired statistical evidence and confidence intervals are reported in \appref{app:robustness}.
\begin{table}[H]
\centering
\captionsetup{skip=4pt}
\caption{\textbf{PTB-XL diagnosis results and patient-matched visual intervention controls.}}\label{tab:clinical_ptbxl}
\small
\renewcommand{\arraystretch}{0.98}
\setlength{\tabcolsep}{5pt}
\begin{tabular}{lrrrr}
\toprule
\rowcolor{softband}\multicolumn{5}{l}{\textbf{A. Diagnosis F1}}\\
\textbf{Diagnosis} & \textbf{ViRe} & \textbf{Medformer} & \textbf{Gain} & \textbf{Patient-bootstrap 95\% CI}\\
\midrule
MI & 72.21 & 71.35 & +0.85 & [-1.42,+3.08]\\
STTC & 74.91 & 73.44 & +1.47 & [-0.99,+3.97]\\
HYP & \textbf{59.49} & 44.17 & \textbf{+15.32} & \textbf{[+10.39,+20.25]}\\
AMI & 72.93 & 71.01 & +1.92 & [-1.23,+5.14]\\
ISCA & 37.07 & 30.15 & +6.92 & [-2.03,+15.94]\\
LVH & \textbf{66.51} & 46.73 & \textbf{+19.78} & \textbf{[+13.98,+25.48]}\\
AFIB & 67.08 & 62.61 & +4.47 & [-2.20,+10.98]\\
\addlinespace[1pt]
\rowcolor{softband}\multicolumn{5}{l}{\textbf{B. Visual intervention controls} \hfill Holm-adjusted $p=0.0008$}\\
\textbf{Intervention} & \multicolumn{2}{c}{\textbf{Macro-F1 drop}} & \multicolumn{2}{c}{\textbf{95\% CI}}\\
\midrule
Zero-mask visual input & \multicolumn{2}{c}{\textbf{-6.19}} & \multicolumn{2}{c}{\textbf{[-8.06,-4.44]}}\\
Cross-patient visual shuffling & \multicolumn{2}{c}{\textbf{-15.61}} & \multicolumn{2}{c}{\textbf{[-17.85,-13.28]}}\\
\bottomrule
\end{tabular}
\end{table}

\textbf{Concept decoding from the frozen vision prior.} On \textbf{MEETI}~\citep{zhang2026meeti}, we probe the frozen visual representation with lightweight concept heads and obtain positive confidence intervals for all 22 supported attributes. For continuous attributes, $\rho$ denotes Spearman's rank correlation coefficient; prolonged QT is evaluated by AUROC. \tabref{tab:meeti_main} reports representative amplitude, interval, synchrony, and QT results, and \tabref{tab:meeti_appendix} summarizes the remaining representation analyses.
\begin{table}[H]
\centering
\caption{\textbf{Representative MEETI concept-decoding results}~\citep{zhang2026meeti}.}\vspace{-3mm}
\label{tab:meeti_main}
\small
\renewcommand{\arraystretch}{1.06}
\setlength{\tabcolsep}{0pt}
\begin{tabular*}{\textwidth}{@{\extracolsep{\fill}} l c r r r @{}}
\toprule
\textbf{Attribute} & \textbf{Metric} & \textbf{CLIP} & \textbf{Random} & \textbf{Gain}\\
\midrule
Amplitude range & $\rho$ & \textbf{0.903} & 0.501 & \textbf{+0.402}\\
PR interval & $\rho$ & \textbf{0.633} & 0.248 & \textbf{+0.384}\\
Cross-lead synchrony & $\rho$ & \textbf{0.684} & 0.320 & \textbf{+0.364}\\
Prolonged QT & AUROC & \textbf{0.904} & 0.656 & \textbf{+0.248}\\
\bottomrule
\end{tabular*}
\end{table}

\textbf{Morphology-aware representation analysis.} Because \textbf{MEETI} pairs ECG waveforms with images, measurements, and clinical reports, we further evaluate the frozen visual representation without retraining, testing alignment with clinical text, whether same-patient representation changes follow measured waveform changes, and whether nearby visual embeddings share morphology-related attributes. As summarized in \tabref{tab:meeti_appendix}, all 12 text directions improve, all 5 longitudinal measurements show positive gains in direction accuracy, and all 21 retrieval attributes have positive confidence intervals, even though the visual encoder is never trained on the benchmark labels.
\begin{table}[H]
\centering
\captionsetup{skip=4pt}
\caption{\textbf{Morphology-aware representation analyses on MEETI.}}
\label{tab:meeti_appendix}
\fontsize{7.5}{8.5}\selectfont
\renewcommand{\arraystretch}{1.10}
\setlength{\tabcolsep}{0pt}
\begin{tabularx}{\textwidth}{@{} >{\raggedright\arraybackslash}p{0.30\textwidth} @{\hspace{4pt}} >{\raggedright\arraybackslash}X @{}}
\toprule
\rowcolor{softband}\textbf{Zero-shot text alignment} & \textbf{12/12 directions}\\
\textbf{Key results} & AF AUROC \textbf{+0.423}; QTc $\rho$ \textbf{+0.284}; amplitude \textbf{+0.239}; synchrony \textbf{+0.232}.\\
\rowcolor{softband}\textbf{Same-patient natural changes} & \textbf{5/5 measurements}\\
\textbf{Key results} & Direction accuracy: HR \textbf{+0.128}; PR \textbf{+0.154}; QRS \textbf{+0.127}; QT \textbf{+0.238}; QTc \textbf{+0.144}.\\
\rowcolor{softband}\textbf{Raw-space retrieval} & \textbf{21/21 positive CIs}\\
\textbf{Key results} & Bradycardia \textbf{+0.337}; HR \textbf{+0.331}; amplitude \textbf{+0.262}; QT/QRS \textbf{+0.256/+0.245}.\\
\bottomrule
\end{tabularx}
\end{table}

The interventions in \tabref{tab:clinical_ptbxl} and the probes in \tabref{tab:meeti_main} and \tabref{tab:meeti_appendix} are complementary: replacing the visual evidence degrades diagnosis, whereas the frozen visual space already organizes recognizable waveform attributes, so the visual branch acts as a morphology-aware retrieval prior rather than a diagnostic shortcut that bypasses the numerical evidence altogether.

\section{Conclusion}
We introduce ViRe, a Vision-Informed Retrieval framework that complements numerical MedTS modeling with a morphology-aware prior from a frozen CLIP vision encoder, whose global \textit{Vision Query} retrieves temporal and channel evidence by cross-attention. Across six public EEG and ECG benchmarks, ViRe achieves the best overall performance on five datasets and improves the strongest baseline, Medformer, by \textbf{6.42\%} on average and by \textbf{16.37\%} on APAVA.

Mechanism analyses favor vision-informed queries and CLIP-Vision over content-free queries and generic image pretraining, and PTB-XL diagnosis results and MEETI concept decoding link the visual prior to recognizable waveform structure. ViRe thus lets waveform morphology guide numerical feature selection, and we hope it encourages further use of frozen visual priors in clinical time series.

\begin{ack}
This work was partially supported by the Research Grants Council (RGC) of Hong Kong under the Collaborative Research Fund (CRF) (No. C5055-24G), the Start-up Fund of The Hong Kong Polytechnic University (No. P0045999), the Seed Fund of the Research Institute for Smart Ageing (No. P0050946), the Tsinghua-PolyU Joint Research Initiative Fund (No. P0056509), and the University Grants Committee (UGC) funding of The Hong Kong Polytechnic University (No. P0053716).
\end{ack}

\clearpage
\setlength{\bibsep}{\ViReBibSep}
\bibliography{ViRe}
\bibliographystyle{unsrtnat}
\appendix
\numberwithin{equation}{section}
\raggedbottom  

\clearpage
\section{Relationship to Numerical-Only Modeling}
\label{app:vision}
\paragraph{Setup.}
For a training set $\{(X_i,Y_i)\}_{i=1}^n$, ViRe renders $X\in\mathbb{R}^{T\times C}$ as $I=\mathcal V(X)$, extracts the frozen CLIP feature $Z=f_{\mathrm{CLIP}}(I)$, and projects it to the global Vision Query $Q=ZW_z+b_z\in\mathbb{R}^{1\times D}$. The numerical backbone produces temporal and channel tokens $\bar U\in\mathbb{R}^{P\times D}$ and $\bar V\in\mathbb{R}^{C\times D}$. ViRe computes $\widetilde U=\mathrm{CrossAttn}(Q,\bar U,\bar U)$ and $\widetilde V=\mathrm{CrossAttn}(Q,\bar V,\bar V)$, followed by $\widehat Y=(\widetilde U+\widetilde V)W_y+b_y$. The numerical counterpart uses mean pooling, $\widehat Y_{\mathrm{Num}}=(\mu(\bar U)+\mu(\bar V))W_y+b_y$.

\paragraph{Lemma 1 (Cross-attention contains mean pooling).}
For standard scaled dot-product attention, there exists a parameter setting such that, for any $Q\in\mathbb{R}^{1\times D}$ and $K\in\mathbb{R}^{S\times D}$,
\begin{equation}
\mathrm{CrossAttn}(Q,K,K)=\mu(K).
\end{equation}
\eqparbreak
\emph{Proof.} For one head, set $W_Q=0$ (equivalently $W_K=0$). Every attention logit is zero, the softmax becomes uniform, and the head returns $S^{-1}\sum_{s=1}^{S}k_sW_V$. After concatenating heads and applying the output projection, the remaining linear map of $\mu(K)$ is absorbed into $(W_y,b_y)$. \hfill$\square$

\paragraph{Theorem 1 (ViRe contains mean-pooled numerical modeling).}
Let $\mathcal F_{\mathrm{ViRe}}$ and $\mathcal F_{\mathrm{Num}}$ denote the ViRe and numerical hypothesis classes. For empirical cross-entropy risk $\widehat{\mathcal L}$, the inclusion $\mathcal F_{\mathrm{Num}}\subseteq\mathcal F_{\mathrm{ViRe}}$ implies
\begin{equation}
\mathcal L^\star_{\mathrm{ViRe}}:=\inf_{f\in\mathcal F_{\mathrm{ViRe}}}\widehat{\mathcal L}(f)
\leq
\inf_{f\in\mathcal F_{\mathrm{Num}}}\widehat{\mathcal L}(f)=:\mathcal L^\star_{\mathrm{Num}}.
\end{equation}
\eqparbreak
\emph{Proof.} Take any $f\in\mathcal F_{\mathrm{Num}}$. By Lemma 1, choose the two retrieval blocks so that $\widetilde U=\mu(\bar U)$ and $\widetilde V=\mu(\bar V)$ for every input, while retaining the same numerical backbone and classifier. ViRe exactly reproduces $f$, establishing both the hypothesis-class inclusion and empirical-risk inequality. \hfill$\square$

\paragraph{Proposition 1 (Complementary structure after numerical compression).}
Let $S=(\bar U,\bar V)$ and let $\bar Z$ denote the projected visual representation. Both are deterministic representations of $X$ and can preserve different aspects of its structure. Positive conditional information, $I(Y;\bar Z\mid S)>0$, yields
\begin{equation}
R^\star(S,\bar Z)=H(Y\mid S,\bar Z)<H(Y\mid S)=R^\star(S).
\end{equation}
\eqparbreak
\emph{Proof.} Under log-loss, Bayes risk equals conditional entropy, and $H(Y\mid S)-H(Y\mid S,\bar Z)=I(Y;\bar Z\mid S)>0$ gives the result. \hfill$\square$

\paragraph{Proposition 2 (Vision-conditioned retrieval realizes adaptive pooling).}
For a token set $K=[k_1,\ldots,k_S]^\top$, one retrieval head produces
\begin{equation}
\mathrm{Attn}(Q,K)=\sum_{s=1}^{S}\alpha_s(Q,K)k_sW_V,\qquad
\alpha(Q,K)=\mathrm{softmax}\!\left(\frac{QW_Q(KW_K)^\top}{\sqrt d}\right).
\end{equation}
\eqparbreak
The weights therefore form a sample-dependent pooling rule indexed jointly by the Vision Query and the numerical tokens. Uniform pooling is recovered by Lemma 1, whereas non-uniform logits yield selective aggregation over temporal or channel evidence. Applying this operator to $\bar U$ and $\bar V$ gives ViRe two complementary adaptive retrieval paths over the original numerical representation.

\emph{Proof.} Lemma 1 gives the uniform-weight member of the family. Whenever two projected key scores differ, the softmax assigns distinct weights to their tokens. Because the scores depend jointly on $Q$ and $K$, changing either the visual query or the numerical tokens changes the pooling coefficients continuously. The operator therefore spans both uniform and sample-adaptive aggregation. \hfill$\square$

\paragraph{Dual-path aggregation.}
The temporal path applies this adaptive weighting over $P$ temporal tokens, emphasizing time-localized morphology such as sharp transitions, recurrent complexes, and long-range waveform structure. The channel path applies the same principle over $C$ channel tokens, emphasizing cross-channel patterns and channel-specific morphology. Their summed representation combines these two views before classification, linking the global vision prior to the original numerical evidence at both temporal and channel resolutions of the input signal.

\clearpage
\section{Data Preprocessing and Train-validation-test Split}
\label{app:data}
We utilize \textbf{three} EEG (APAVA, ADFTD, and TDBrain) and \textbf{three} ECG (PTB, PTB-XL, and MIMIC) datasets under the Subject-Independent setting~\citep{wang2024subin0}, where samples from the same subject are exclusively divided into training, validation, or test sets. The data preprocessing and train-validation-test split protocol follows Medformer~\citep{wang2024medformer}. Since MIMIC is not included in the Medformer benchmark, we download and preprocess it with the identical protocol to keep the comparison fair.

\textbf{APAVA.} The Alzheimer’s Patients’ Relatives Association of Valladolid (APAVA) dataset~\citep{escudero2006apava} is a public two-class EEG benchmark containing recordings from 23 subjects, i.e., 12 Alzheimer’s disease (AD) patients and 11 healthy controls (HC). For each trial, we extract 9 half-overlapping windows, where each window corresponds to a 1-second sequence with 256 timestamps. In total, this yields 5,967 samples. We reserve subjects 15,16,19,20 for validation and 1,2,17,18 for testing, and use the remaining subjects, together with all of their samples, for training.

\textbf{ADFTD.} The Alzheimer’s Disease and FronTotemporal Dementia (ADFTD) dataset~\citep{miltiadous2023adftd} is a public EEG dataset with three classes recorded over 19 channels, including 36 AD patients, 23 Frontotemporal Dementia (FTD) patients, and 29 HCs. We first resample each trial from 500 Hz to 256 Hz, then slice it into non-overlapping 1-second windows with 256 timestamps, discarding any trailing segments shorter than 1 second. This yields 69,752 samples. We perform a subject-wise split, allocating 60\%, 20\%, and 20\% of subjects (and all corresponding samples) to the training, validation, and test sets, respectively, so that no subject appears in more than one partition.

\textbf{TDBrain.} TDBrain~\citep{van2022tdbrain} is a large permissioned EEG dataset with 33 channels collected from 1,274 individuals. Each subject provides two recordings (eyes-open and eyes-closed). We consider a balanced subset consisting of 25 Parkinson’s disease (PD) subjects and 25 HCs, using only the eyes-closed recordings. Each trial is partitioned into non-overlapping 1-second segments (256 timestamps), and segments shorter than 1 second are removed. This results in 6,240 samples. We assign subjects 18,19,20,21,46,47,48,49 to the validation set and 22,23,24,25,50,51,52,53 to the test set; all remaining subjects, together with all of their segments, are used exclusively for training.

\textbf{PTB.} The PTB dataset~\citep{physiobank2000ptb} is a public ECG dataset collected from 290 subjects with 15 leads and 8 labels (7 cardiac conditions plus healthy control). In this work, we select 198 subjects from the myocardial infarction and HC categories. We downsample the original 500 Hz recordings to 250 Hz and standardize the signals. We then convert each recording into single-heartbeat samples: R-peaks are detected across all leads, outlier intervals are filtered out, and each beat is extracted around its R-peak. To enforce a fixed length, we zero-pad shorter beats using the maximum beat duration across all channels as the reference. This procedure produces 64,356 heartbeat samples, which are split subject-wise into training, validation, and test sets at 60\% : 20\% : 20\%.

\textbf{PTB-XL.} PTB-XL~\citep{wagner2020ptb-xl} is a large-scale public ECG dataset with 12 leads from 18,869 subjects and 5 diagnostic categories (4 diseases plus healthy control). To avoid label inconsistency, we remove subjects whose diagnoses differ across trials, leaving 17,596 subjects. Each record spans 10 seconds and is available at 100 Hz and 500 Hz; we use the 500 Hz version, resample it to 250 Hz, and apply standard scaling. Next, we segment each trial into non-overlapping 1-second windows (250 timestamps) and discard any remainder shorter than 1 second. This yields 191,400 samples. We use a subject-level 60\% : 20\% : 20\% split, keeping all windows from each subject together.

\textbf{MIMIC.} MIMIC~\citep{gow2023mimic} is an ECG dataset with binary labels indicating heart disease versus healthy control. Each recording is 10 seconds long and sampled at 500 Hz. We curate a subset of patients with consistent diagnostic labels across records, then downsample signals to 250 Hz and standardize them. R-peaks are detected on all leads to segment each record into individual heartbeats, and noisy or outlier beats are removed. Each heartbeat is aligned to its R-peak and zero-padded to a uniform length determined by the maximum beat duration observed in the dataset. This preprocessing yields 204,370 samples. Finally, we split subjects (and all corresponding heartbeats) into training, validation, and test sets using a 60\% : 20\% : 20\% subject-level split, keeping all heartbeats of a subject together.

\textbf{Summary.} All six datasets follow the same pipeline: subject-level partitioning is fixed before preprocessing, each subject contributes to exactly one split, and the numerical input and its rendered waveform image are derived from the same preprocessed segment. The visual branch thus never observes information unavailable to the numerical branch, and \tabref{tab:data_inf} describes both modalities.

\clearpage
\section{Data Augmentation}
\label{app:augmentation}
During training, each input is augmented by exactly one augmentation operation, which is sampled uniformly at random from the six transformations summarized in \tabref{tab:data_augmentation}.

\begin{table}[!htb]
\centering
\captionsetup{skip=3pt}
\caption{Training-time augmentations and default settings.}
\label{tab:data_augmentation}
\small
\setlength{\tabcolsep}{6pt}
\renewcommand{\arraystretch}{\ViReAugmentationStretch}
\begin{tabularx}{\textwidth}{>{\raggedright\arraybackslash}p{0.22\textwidth}>{\raggedright\arraybackslash}X>{\raggedright\arraybackslash}p{0.17\textwidth}}
\toprule
\textbf{Operation} & \textbf{Transformation} & \textbf{Default} \\
\midrule
Temporal flipping & Reverse the sequence along the time axis. & $\textit{prob}=0.5$ \\
Channel shuffling & Randomly permute the channel order. & $\textit{prob}=0.5$ \\
Temporal masking & Mask timestamps shared across all channels. & $\textit{ratio}=0.1$ \\
Frequency masking & Suppress randomly selected frequency bands and transform the signal back to the time domain. & $\textit{ratio}=0.1$ \\
Jittering & Add random noise sampled from $[0,1]$ and scale its magnitude. & $\textit{scale}=0.1$ \\
Dropout & Randomly set a fraction of signal values to zero. & $\textit{ratio}=0.1$ \\
\bottomrule
\end{tabularx}
\end{table}

For each training instance, we draw one operation $a$ uniformly from the six-operation pool $\mathcal A$. The numerical and visual inputs are
\begin{equation}
X'=a(X),\qquad I'=\mathcal V(X').
\end{equation}
\eqparbreak
Both temporal and channel branches therefore receive the same transformed signal, so masked intervals, channel permutations, and local perturbations stay synchronized.

Only one operation is sampled in each forward pass. This avoids compounding several perturbations into an unrealistic waveform, while keeping the selected transformation and its default strength explicit. The sampling rule and settings in \tabref{tab:data_augmentation} are shared across datasets.

Augmentation is disabled for validation and testing. Every held-out sample is processed by the deterministic preprocessing and rendering pipeline used for model selection and final evaluation.

\paragraph{Rationale.} The pool targets nuisance factors that are common in clinical recordings rather than label-relevant morphology. Temporal flipping and channel shuffling discourage the encoders from memorizing absolute positions or a fixed electrode order; temporal masking and dropout imitate transient electrode dropout and missing samples; frequency masking removes narrow bands in the way that filtering or line-noise suppression does; and jittering models low-amplitude sensor noise. Because the rendering is regenerated from the augmented signal, the visual branch is exposed to the same nuisance variation and cannot rely on rendering-specific artifacts.

\clearpage
\section{Implementation Details}
\label{app:implementation}
\subsection{Implementation Details of All Baselines}
\label{app:baseline_implementation}
All baseline methods are implemented on top of Medformer~\citep{wang2024medformer}, which unifies competing approaches within a shared training pipeline, enabling a consistent comparison. We benchmark ten Transformer baselines in this common pipeline: Autoformer~\citep{wu2021autoformer}, FEDformer~\citep{zhou2022fedformer}, Informer~\citep{zhou2021informer}, iTransformer~\citep{liu2023itransformer}, MTST~\citep{zhang2024mtst}, Nonformer~\citep{liu2022nonfromer}, PatchTST~\citep{nie2022patchtst}, Reformer~\citep{kitaev2019reformer}, Medformer~\citep{wang2024medformer}, and the vanilla Transformer~\citep{vaswani2017attention}, all trained and evaluated under identical settings.

For Medformer, we reproduce the reported results using the authors' official implementation. For the remaining baselines, we standardize the architecture by using a 6-layer encoder, setting the attention embedding dimension $D$ to 128, and the hidden size of the feed-forward network to 256. We train all models with the Adam optimizer using a learning rate of $1\mathrm{e}{-4}$. The batch size is fixed to $\{32,32,128,128,128,128\}$ for APAVA, TDBrain, ADFTD, PTB, PTB-XL, and MIMIC, respectively. Each model is trained for 100 epochs with early stopping (patience $=10$) based on validation macro F1-Score. We checkpoint the model achieving the best validation F1-Score and report test performance accordingly. We evaluate Accuracy, macro-Precision, macro-Recall, macro-F1, macro-AUROC, and macro-AUPRC. All experiments are repeated with five random seeds under fixed train, validation, and test splits, and results are reported as mean$\pm$std.

\textbf{Autoformer.} Autoformer~\citep{wu2021autoformer} replaces standard self-attention with an auto-correlation operator tailored for time series forecasting It further incorporates a decomposition module that separates the input into trend/cyclical and seasonal components to facilitate representation learning.

\textbf{FEDformer.} FEDformer~\citep{zhou2022fedformer} exploits Fourier-domain representations through frequency-enhanced blocks and frequency-domain attention, and it introduces a decomposition mechanism that replaces layer normalization in the Transformer to improve its modeling capacity.

\textbf{Informer.} Informer~\citep{zhou2021informer} introduces ProbSparse attention and a one-shot generative forecasting paradigm to reduce both the computational and memory costs of long-sequence modeling.

\textbf{iTransformer.} iTransformer~\citep{liu2023itransformer} forms tokens by embedding entire channels and correspondingly swaps dimensions in normalization and feed-forward modules.

\textbf{MTST.} MTST~\citep{zhang2024mtst} uses multi-scale tokens with heterogeneous patch lengths.

\textbf{Nonformer.} Nonformer~\citep{liu2022nonfromer} models non-stationary dynamics with de-stationary attention and applies paired normalization and denormalization to mitigate over-stationarization.

\textbf{PatchTST.} PatchTST~\citep{nie2022patchtst} constructs patch tokens from single-channel temporal segments, enlarging the receptive field of each token and improving long-horizon temporal modeling.

\textbf{Reformer.} Reformer~\citep{kitaev2019reformer} approximates dot-product attention with locality-sensitive hashing and uses reversible residual layers to reduce attention complexity and training memory.

\textbf{Transformer.} The vanilla Transformer~\citep{vaswani2017attention}, introduced in ``Attention Is All You Need,'' can be adapted to time series by treating each multivariate timestamp as a token.

\textbf{Medformer.} Medformer~\citep{wang2024medformer} is a multi-granularity patching Transformer for medical time series classification, capturing local and long-range dependencies.

\paragraph{Common evaluation protocol.} All methods use the same subject-disjoint partitions, preprocessing, label definitions, validation-based early stopping, six evaluation metrics, and five random seeds. The test set is evaluated only after the best validation checkpoint has been selected.

\paragraph{Input controls.} Each baseline receives the same numerical input under the partitions used by ViRe. Architecture-specific modules follow the corresponding released implementations, while dataset statistics and evaluation settings remain fixed across methods.

\newpage
\subsection{Implementation Details of the Proposed ViRe}
ViRe uses shared settings across datasets, summarized in \tabref{tab:implementation_details}. Code and training scripts are publicly available in the \href{\ViReRepoURL}{GitHub Repo}, together with the executable visualization notebook and the precomputed CLIP features of all six benchmark datasets used in this paper.
\begin{table}[!htb]
\centering
\captionsetup{skip=4pt}
\caption{\textbf{Implementation configuration of ViRe.}}
\label{tab:implementation_details}
\small
\setlength{\tabcolsep}{6pt}
\renewcommand{\arraystretch}{\ViReImplementationStretch}
\resizebox{\textwidth}{!}{%
\begin{tabular}{p{0.30\textwidth}p{0.60\textwidth}}
\toprule
\textbf{Item} & \textbf{Setting} \\
\midrule
Batch size & $B=128$ \\
Learning rate & $1\mathrm{e}{-4}$ \\
Model dimension & $D=128$ \\
Temporal Encoder depth & $M=6$ for all datasets \\
Channel Encoder depth & $N=6$ by default; $N=0$ for TDBrain \\
Temporal granularity $L$ & Dataset order: APAVA, TDBrain, ADFTD, PTB, PTB-XL, MIMIC; $L=\{1,3,8,1,6,6\}$ \\
\bottomrule
\end{tabular}%
}
\end{table}

\paragraph{Training protocol.} ViRe is optimized with Adam at a learning rate of $1\mathrm{e}{-4}$ for up to 100 epochs, with early stopping (patience $=10$) on validation macro-F1. We use five random seeds under fixed subject-disjoint splits and report mean$\pm$std. The CLIP encoder remains frozen throughout optimization, so the vision branch introduces no additional trainable backbone parameters.

\paragraph{Dataset-specific configuration.} The model dimension is fixed at $D=128$ and the temporal encoder depth at $M=6$. The channel depth and temporal granularity $L$ follow \tabref{tab:implementation_details}; all remaining optimization and evaluation settings are shared across datasets.

\paragraph{Visual feature caching.} Waveform rendering is deterministic. During training, the frozen CLIP features are cached once per sample, so repeated optimization does not rerun the visual encoder. This is also the execution mode used for the training-cost measurements in \appref{app:computational_overhead}.

\paragraph{Trainable components.} Optimization updates the numerical encoders, dimension-alignment layer, two retrieval blocks, and classifier. The CLIP parameters remain fixed in every experiment.

\paragraph{Checkpoint evaluation.} The best validation macro-F1 checkpoint is evaluated on the test split for each seed, using the same macro-averaged metrics as the baselines in \appref{app:baseline_implementation}.

\paragraph{Vision branch.} The frozen CLIP Vision Encoder is used in inference mode: images are produced by the operator in \appref{app:visualization_operator}, encoded once, and the resulting $\tilde D$-dimensional features are cached. Only the Dimension Align layer that maps them to $D=128$ is trained, so the visual branch adds only one linear projection to the total trainable parameter count.

\paragraph{Retrieval cost.} Because a single Vision Query attends over $P$ temporal and $C$ channel tokens, each retrieval block costs $O((P+C)D)$ per sample, which is negligible compared with the quadratic self-attention cost of the two Transformer Encoders over $P$ and $C$ tokens.

\subsection{Visualization Operator}
\label{app:visualization_operator}
\begin{figure}[!htb]
\centering
\makebox[\textwidth][c]{\includegraphics[width=\textwidth]{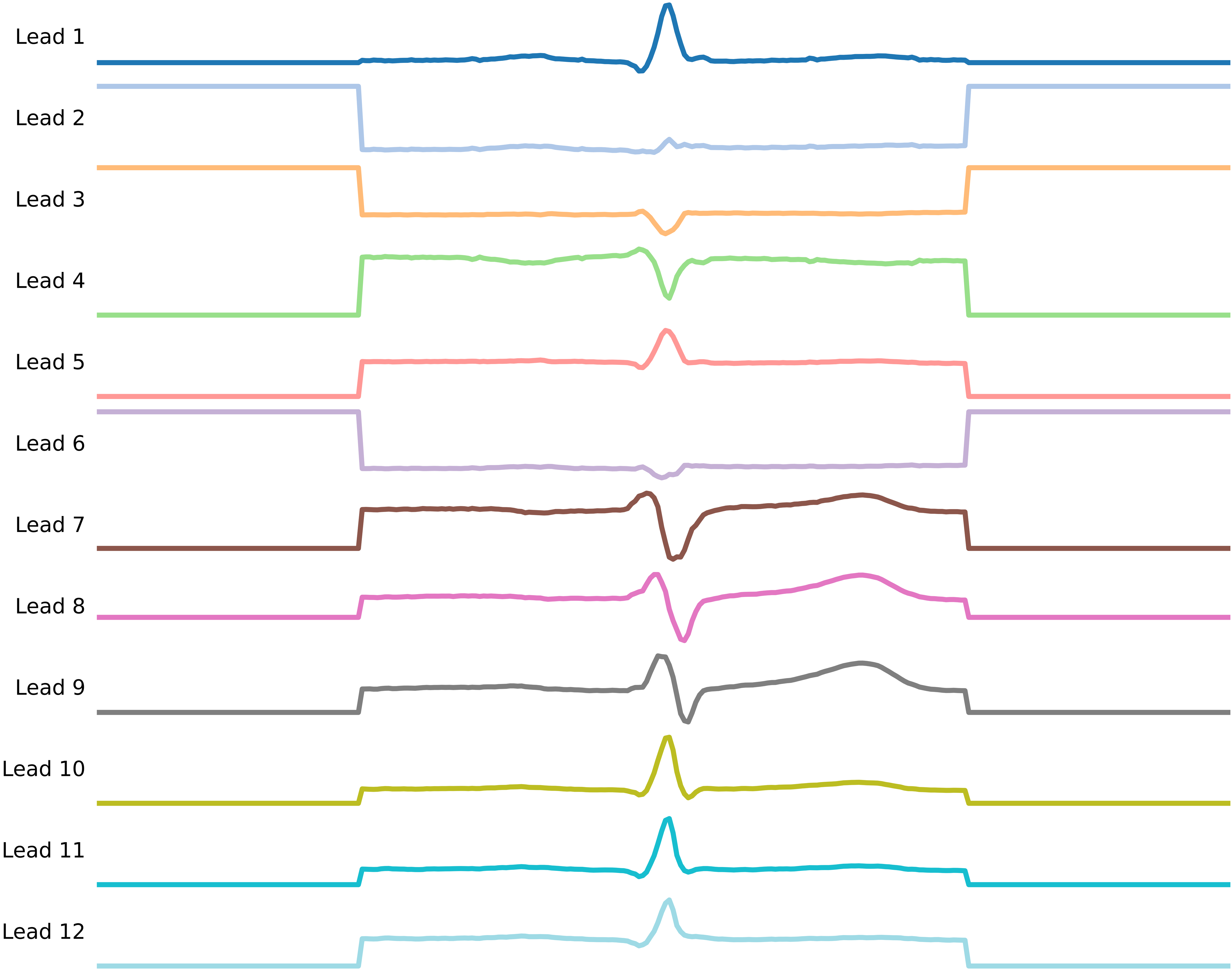}}
\caption{Visualization example. Stacked twelve-lead PTB waveform rendered by $\mathcal V(\cdot)$.}
\label{fig:visual_example}
\end{figure}

ViRe deterministically transforms each preprocessed multichannel MedTS sample into a CLIP-readable stacked waveform image. Each channel is rendered on a separate panel with fixed axes, after which the panels are vertically concatenated. This standardized layout preserves channel morphology and cross-channel timing while removing decorative plot elements. \algref{alg:visualization_operator} specifies the operator, and an executable version is available as a notebook in the \href{\ViReNotebookURL}{GitHub Repo}. The same operator is applied without modification to training, validation, and test samples, and \figref{fig:visual_example} shows the rendering of a twelve-lead PTB sample produced by this operator with all decorations removed.

\begin{algorithm}[!htb]
\caption{Visualization Operator $\mathcal V(\cdot)$}
\label{alg:visualization_operator}
\small
\begin{algorithmic}[1]
\Require Preprocessed MedTS sample $X\in\mathbb{R}^{C\times T}$, sampling rate $f_s$
\Ensure Stacked RGB waveform image $I_{\mathrm{stack}}$
\State Construct $t\gets[0,1,\ldots,T-1]/f_s$ and initialize $\mathcal P\gets[\ ]$.
\For{$c=1$ to $C$}
    \State Render $X_{c,:}$ against $t$ on an independent canvas.
    \State Remove ticks, spines, legends, grids, and axis decorations.
    \State Export RGB panel $P_c$ and append it to $\mathcal P$.
\EndFor
\State $I_{\mathrm{stack}}\gets\mathrm{VConcat}(P_1,\ldots,P_C)$.
\State \Return $I_{\mathrm{stack}}$.
\end{algorithmic}
\end{algorithm}

\clearpage
\section{Full Classification Results}
\label{app:full_results}
\begin{center}
\captionsetup{type=table,font=small,skip=4pt}
\captionof{table}{\textbf{Full subject-independent classification results.} Mean$\pm$std over five seeds; best is bolded and second-best is shown in blue italics.}
\label{tab:full_results}
\begingroup
\renewcommand{\arraystretch}{\ViReFullResultsStretch}
\resizebox{\ViReFullResultsWidth}{!}{%
\begin{tabular}{clccccccc}

    \toprule
    \scalebox{1.2}{\textbf{Datasets}} & \scalebox{1.2}{\textbf{Models}} & \multicolumn{1}{c}{\scalebox{1.2}{\textbf{Accuracy}}} & \multicolumn{1}{c}{\scalebox{1.2}{\textbf{Precision}}} & \multicolumn{1}{c}{\scalebox{1.2}{\textbf{Recall}}} & \multicolumn{1}{c}{\scalebox{1.2}{\textbf{F1-Score}}} & \multicolumn{1}{c}{\scalebox{1.2}{\textbf{AUROC}}} & \multicolumn{1}{c}{\scalebox{1.2}{\textbf{AUPRC}}} & \multicolumn{1}{c}{\scalebox{1.2}{\textbf{Avg}}} \\

    \midrule
    \multirow{11}{*}{\begin{tabular}[c]{@{}c@{}}\;\makecell{\scalebox{1.2}{\textbf{APAVA}} \\ \scalebox{1.2}{(2-Classes)} \\ \scalebox{1.2}{(EEG)} } \end{tabular}} 
    & \scalebox{1.2}{\textbf{Autoformer}}     & \scalebox{1.3}{68.64\std{1.82}} & \scalebox{1.3}{68.48\std{2.10}} & \scalebox{1.3}{68.77\std{2.27}} & \scalebox{1.3}{68.06\std{1.94}} & \scalebox{1.3}{75.94\std{3.61}} & \scalebox{1.3}{74.38\std{4.05}} & \scalebox{1.3}{70.71\std{2.63}} \\
    & \scalebox{1.2}{\textbf{FEDformer}}      & \scalebox{1.3}{74.94\std{2.15}} & \scalebox{1.3}{74.59\std{1.50}} & \scalebox{1.3}{73.56\std{3.55}} & \scalebox{1.3}{73.51\std{3.39}} & \scalebox{1.3}{83.72\std{1.97}} & \scalebox{1.3}{82.94\std{2.37}} & \scalebox{1.3}{77.21\std{2.49}} \\
    & \scalebox{1.2}{\textbf{Informer}}       & \scalebox{1.3}{73.11\std{4.40}} & \scalebox{1.3}{75.17\std{6.06}} & \scalebox{1.3}{69.17\std{4.56}} & \scalebox{1.3}{69.47\std{5.06}} & \scalebox{1.3}{70.46\std{4.91}} & \scalebox{1.3}{70.75\std{5.27}} & \scalebox{1.3}{71.36\std{5.04}} \\
    & \scalebox{1.2}{\textbf{iTransformer}}   & \scalebox{1.3}{74.55\std{1.66}} & \scalebox{1.3}{74.77\std{2.10}} & \scalebox{1.3}{71.76\std{1.72}} & \scalebox{1.3}{72.30\std{1.79}} & \scalebox{1.3}{\second{85.59\std{1.55}}} & \scalebox{1.3}{\second{84.39\std{1.57}}} & \scalebox{1.3}{77.23\std{1.73}} \\
    & \scalebox{1.2}{\textbf{MTST}}           & \scalebox{1.3}{71.14\std{1.59}} & \scalebox{1.3}{79.30\std{0.97}} & \scalebox{1.3}{65.27\std{2.28}} & \scalebox{1.3}{64.01\std{3.16}} & \scalebox{1.3}{68.87\std{2.34}} & \scalebox{1.3}{71.06\std{1.60}} & \scalebox{1.3}{69.94\std{1.99}} \\
    & \scalebox{1.2}{\textbf{Nonformer}}      & \scalebox{1.3}{71.89\std{3.81}} & \scalebox{1.3}{71.80\std{4.58}} & \scalebox{1.3}{69.44\std{3.56}} & \scalebox{1.3}{69.74\std{3.84}} & \scalebox{1.3}{70.55\std{2.96}} & \scalebox{1.3}{70.78\std{4.08}} & \scalebox{1.3}{70.70\std{3.81}} \\
    & \scalebox{1.2}{\textbf{PatchTST}}       & \scalebox{1.3}{67.03\std{1.65}} & \scalebox{1.3}{78.76\std{1.28}} & \scalebox{1.3}{59.91\std{2.02}} & \scalebox{1.3}{55.97\std{3.10}} & \scalebox{1.3}{65.65\std{0.28}} & \scalebox{1.3}{67.99\std{0.76}} & \scalebox{1.3}{65.89\std{1.52}} \\
    & \scalebox{1.2}{\textbf{Reformer}}       & \scalebox{1.3}{78.70\std{2.00}} & \scalebox{1.3}{\second{82.50\std{3.95}}} & \scalebox{1.3}{75.00\std{1.61}} & \scalebox{1.3}{75.93\std{1.82}} & \scalebox{1.3}{73.94\std{1.40}} & \scalebox{1.3}{76.04\std{1.14}} & \scalebox{1.3}{77.02\std{1.99}} \\
    & \scalebox{1.2}{\textbf{Transformer}}    & \scalebox{1.3}{76.30\std{4.72}} & \scalebox{1.3}{77.64\std{5.95}} & \scalebox{1.3}{73.09\std{5.01}} & \scalebox{1.3}{73.75\std{5.38}} & \scalebox{1.3}{72.50\std{6.60}} & \scalebox{1.3}{73.23\std{7.60}} & \scalebox{1.3}{74.42\std{5.88}} \\
    & \scalebox{1.2}{\textbf{Medformer}}      & \scalebox{1.3}{\second{78.74\std{0.64}}} & \scalebox{1.3}{81.11\std{0.84}} & \scalebox{1.3}{\second{75.40\std{0.66}}} & \scalebox{1.3}{\second{76.31\std{0.71}}} & \scalebox{1.3}{83.20\std{0.91}} & \scalebox{1.3}{83.66\std{0.92}} & \scalebox{1.3}{\second{79.74\std{0.78}}} \\
    
    \rowcolor{purple!22}& \scalebox{1.2}{\textbf{ViRe}}  & \scalebox{1.3}{\first{91.43\std{1.03}}} & \scalebox{1.3}{\first{91.07\std{1.16}}} & \scalebox{1.3}{\first{91.40\std{0.79}}} & \scalebox{1.3}{\first{91.19\std{1.01}}} & \scalebox{1.3}{\first{95.99\std{0.34}}} & \scalebox{1.3}{\first{95.67\std{0.44}}} & \scalebox{1.3}{\first{92.79\std{0.80}}}\\

    \midrule
    \multirow{11}{*}{\begin{tabular}[c]{@{}c@{}}\;\makecell{\scalebox{1.2}{\textbf{ADFTD}} \\ \scalebox{1.2}{(3-Classes)}  \\ \scalebox{1.2}{(EEG)}} \end{tabular}} 
    & \scalebox{1.2}{\textbf{Autoformer}}     & \scalebox{1.3}{45.25\std{1.48}} & \scalebox{1.3}{43.67\std{1.94}} & \scalebox{1.3}{42.96\std{2.03}} & \scalebox{1.3}{42.59\std{1.85}} & \scalebox{1.3}{61.02\std{1.82}} & \scalebox{1.3}{43.10\std{2.30}} & \scalebox{1.3}{46.43\std{1.90}} \\
    & \scalebox{1.2}{\textbf{FEDformer}}      & \scalebox{1.3}{46.30\std{0.59}} & \scalebox{1.3}{46.05\std{0.76}} & \scalebox{1.3}{44.22\std{1.38}} & \scalebox{1.3}{43.91\std{1.37}} & \scalebox{1.3}{62.62\std{1.75}} & \scalebox{1.3}{46.11\std{1.44}} & \scalebox{1.3}{48.20\std{1.22}} \\
    & \scalebox{1.2}{\textbf{Informer}}       & \scalebox{1.3}{48.45\std{1.96}} & \scalebox{1.3}{46.54\std{1.68}} & \scalebox{1.3}{46.06\std{1.84}} & \scalebox{1.3}{45.74\std{1.38}} & \scalebox{1.3}{65.87\std{1.27}} & \scalebox{1.3}{47.60\std{1.30}} & \scalebox{1.3}{50.04\std{1.57}} \\
    & \scalebox{1.2}{\textbf{iTransformer}}   & \scalebox{1.3}{52.60\std{1.59}} & \scalebox{1.3}{46.79\std{1.27}} & \scalebox{1.3}{47.28\std{1.29}} & \scalebox{1.3}{46.79\std{1.13}} & \scalebox{1.3}{67.26\std{1.16}} & \scalebox{1.3}{49.53\std{1.21}} & \scalebox{1.3}{51.71\std{1.28}} \\
    & \scalebox{1.2}{\textbf{MTST}}           & \scalebox{1.3}{45.60\std{2.03}} & \scalebox{1.3}{44.70\std{1.33}} & \scalebox{1.3}{45.05\std{1.30}} & \scalebox{1.3}{44.31\std{1.74}} & \scalebox{1.3}{62.50\std{0.81}} & \scalebox{1.3}{45.16\std{0.85}} & \scalebox{1.3}{47.89\std{1.34}} \\
    & \scalebox{1.2}{\textbf{Nonformer}}      & \scalebox{1.3}{49.95\std{1.05}} & \scalebox{1.3}{47.71\std{0.97}} & \scalebox{1.3}{47.46\std{1.50}} & \scalebox{1.3}{46.96\std{1.35}} & \scalebox{1.3}{66.23\std{1.37}} & \scalebox{1.3}{47.33\std{1.78}} & \scalebox{1.3}{50.94\std{1.34}} \\
    & \scalebox{1.2}{\textbf{PatchTST}}       & \scalebox{1.3}{44.37\std{0.95}} & \scalebox{1.3}{42.40\std{1.13}} & \scalebox{1.3}{42.06\std{1.48}} & \scalebox{1.3}{41.97\std{1.37}} & \scalebox{1.3}{60.08\std{1.50}} & \scalebox{1.3}{42.49\std{1.79}} & \scalebox{1.3}{45.56\std{1.37}} \\
    & \scalebox{1.2}{\textbf{Reformer}}       & \scalebox{1.3}{50.78\std{1.17}} & \scalebox{1.3}{49.64\std{1.49}} & \scalebox{1.3}{{49.89\std{1.67}}} & \scalebox{1.3}{47.94\std{0.69}} & \scalebox{1.3}{{69.17\std{1.58}}} & \scalebox{1.3}{\second{51.73\std{1.94}}} & \scalebox{1.3}{53.19\std{1.42}} \\
    & \scalebox{1.2}{\textbf{Transformer}}    & \scalebox{1.3}{50.47\std{2.14}} & \scalebox{1.3}{49.13\std{1.83}} & \scalebox{1.3}{48.01\std{1.53}} & \scalebox{1.3}{48.09\std{1.59}} & \scalebox{1.3}{67.93\std{1.59}} & \scalebox{1.3}{48.93\std{2.02}} & \scalebox{1.3}{52.09\std{1.78}} \\
    & \scalebox{1.2}{\textbf{Medformer}}      & \scalebox{1.3}{{\second{53.27\std{1.54}}}} & \scalebox{1.3}{\second{51.02\std{1.57}}} & \scalebox{1.3}{\second{50.71\std{1.55}}} & \scalebox{1.3}{\second{50.65\std{1.51}}} & \scalebox{1.3}{\second{70.93\std{1.19}}} & \scalebox{1.3}{51.21\std{1.32}} & \scalebox{1.3}{\second{54.63\std{1.45}}} \\
    
    \rowcolor{purple!22}& \scalebox{1.2}{\textbf{ViRe}}  & \scalebox{1.3}{\first{57.83\std{1.82}}} & \scalebox{1.3}{\first{56.44\std{2.49}}} & \scalebox{1.3}{\first{53.75\std{3.04}}} & \scalebox{1.3}{\first{54.02\std{3.19}}} & \scalebox{1.3}{\first{76.59\std{1.62}}} & \scalebox{1.3}{\first{60.33\std{2.78}}} & \scalebox{1.3}{\first{59.83\std{2.49}}}\\

    \midrule
    \multirow{11}{*}{\begin{tabular}[c]{@{}c@{}}\;\makecell{\scalebox{1.2}{\textbf{TDBrain}} \\ \scalebox{1.2}{(2-Classes)}  \\ \scalebox{1.2}{(EEG)}} \end{tabular}} 
    & \scalebox{1.2}{\textbf{Autoformer}}     & \scalebox{1.3}{87.33\std{3.79}} & \scalebox{1.3}{88.06\std{3.56}} & \scalebox{1.3}{87.33\std{3.79}} & \scalebox{1.3}{87.26\std{3.84}} & \scalebox{1.3}{93.81\std{2.26}} & \scalebox{1.3}{93.32\std{2.42}} & \scalebox{1.3}{89.52\std{3.28}} \\
    & \scalebox{1.2}{\textbf{FEDformer}}      & \scalebox{1.3}{78.13\std{1.98}} & \scalebox{1.3}{78.52\std{1.91}} & \scalebox{1.3}{78.13\std{1.98}} & \scalebox{1.3}{78.04\std{2.01}} & \scalebox{1.3}{86.56\std{1.86}} & \scalebox{1.3}{86.48\std{1.99}} & \scalebox{1.3}{80.98\std{1.96}} \\
    & \scalebox{1.2}{\textbf{Informer}}       & \scalebox{1.3}{89.02\std{2.50}} & \scalebox{1.3}{89.43\std{2.14}} & \scalebox{1.3}{89.02\std{2.50}} & \scalebox{1.3}{88.98\std{2.54}} & \scalebox{1.3}{96.64\std{0.68}} & \scalebox{1.3}{96.75\std{0.63}} & \scalebox{1.3}{91.64\std{1.83}} \\
    & \scalebox{1.2}{\textbf{iTransformer}}   & \scalebox{1.3}{74.67\std{1.06}} & \scalebox{1.3}{74.71\std{1.06}} & \scalebox{1.3}{74.67\std{1.06}} & \scalebox{1.3}{74.65\std{1.06}} & \scalebox{1.3}{83.37\std{1.14}} & \scalebox{1.3}{83.73\std{1.27}} & \scalebox{1.3}{77.63\std{1.11}} \\
    & \scalebox{1.2}{\textbf{MTST}}           & \scalebox{1.3}{76.96\std{3.76}} & \scalebox{1.3}{77.24\std{3.59}} & \scalebox{1.3}{76.96\std{3.76}} & \scalebox{1.3}{76.88\std{3.83}} & \scalebox{1.3}{85.27\std{4.46}} & \scalebox{1.3}{82.81\std{5.64}} & \scalebox{1.3}{79.35\std{4.17}} \\
    & \scalebox{1.2}{\textbf{Nonformer}}      & \scalebox{1.3}{87.88\std{2.48}} & \scalebox{1.3}{88.86\std{1.84}} & \scalebox{1.3}{87.88\std{2.48}} & \scalebox{1.3}{87.78\std{2.56}} & \scalebox{1.3}{\second{97.05\std{0.68}}} & \scalebox{1.3}{\second{96.99\std{0.68}}} & \scalebox{1.3}{91.07\std{1.79}} \\
    & \scalebox{1.2}{\textbf{PatchTST}}       & \scalebox{1.3}{79.25\std{3.79}} & \scalebox{1.3}{79.60\std{4.09}} & \scalebox{1.3}{79.25\std{3.79}} & \scalebox{1.3}{79.20\std{3.77}} & \scalebox{1.3}{87.95\std{4.96}} & \scalebox{1.3}{86.36\std{6.67}} & \scalebox{1.3}{81.94\std{4.51}} \\
    & \scalebox{1.2}{\textbf{Reformer}}       & \scalebox{1.3}{87.92\std{2.01}} & \scalebox{1.3}{88.64\std{1.40}} & \scalebox{1.3}{87.92\std{2.01}} & \scalebox{1.3}{87.85\std{2.08}} & \scalebox{1.3}{96.30\std{0.54}} & \scalebox{1.3}{96.40\std{0.45}} & \scalebox{1.3}{90.84\std{1.42}} \\
    & \scalebox{1.2}{\textbf{Transformer}}    & \scalebox{1.3}{87.17\std{1.67}} & \scalebox{1.3}{87.99\std{1.68}} & \scalebox{1.3}{87.17\std{1.67}} & \scalebox{1.3}{87.10\std{1.68}} & \scalebox{1.3}{96.28\std{0.92}} & \scalebox{1.3}{96.34\std{0.81}} & \scalebox{1.3}{90.34\std{1.41}} \\
    & \scalebox{1.2}{\textbf{Medformer}}      & \scalebox{1.3}{\second{89.62\std{0.81}}} & \scalebox{1.3}{\second{89.68\std{0.78}}} & \scalebox{1.3}{\second{89.62\std{0.81}}} & \scalebox{1.3}{\second{89.62\std{0.81}}} & \scalebox{1.3}{96.41\std{0.35}} & \scalebox{1.3}{96.51\std{0.33}} & \scalebox{1.3}{\second{91.91\std{0.65}}} \\
    
    \rowcolor{purple!22}& \scalebox{1.2}{\textbf{ViRe}}  & \scalebox{1.3}{\first{93.96\std{0.75}}} & \scalebox{1.3}{\first{94.03\std{0.72}}} & \scalebox{1.3}{\first{93.96\std{0.75}}} & \scalebox{1.3}{\first{93.96\std{0.75}}} & \scalebox{1.3}{\first{98.65\std{0.34}}} & \scalebox{1.3}{\first{98.68\std{0.35}}} & \scalebox{1.3}{\first{95.54\std{0.61}}}\\

    \midrule
    \multirow{11}{*}{\begin{tabular}[c]{@{}c@{}}\;\makecell{\scalebox{1.2}{\textbf{PTB}} \\ \scalebox{1.2}{(2-Classes)}  \\ \scalebox{1.2}{(ECG)}} \end{tabular}} 
    & \scalebox{1.2}{\textbf{Autoformer}}     & \scalebox{1.3}{73.35\std{2.10}} & \scalebox{1.3}{72.11\std{2.89}} & \scalebox{1.3}{63.24\std{3.17}} & \scalebox{1.3}{63.69\std{3.84}} & \scalebox{1.3}{78.54\std{3.48}} & \scalebox{1.3}{74.25\std{3.53}} & \scalebox{1.3}{70.86\std{3.17}} \\
    & \scalebox{1.2}{\textbf{FEDformer}}      & \scalebox{1.3}{76.05\std{2.54}} & \scalebox{1.3}{77.58\std{3.61}} & \scalebox{1.3}{66.10\std{3.55}} & \scalebox{1.3}{67.14\std{4.37}} & \scalebox{1.3}{85.93\std{4.31}} & \scalebox{1.3}{82.59\std{5.42}} & \scalebox{1.3}{75.90\std{3.97}} \\
    & \scalebox{1.2}{\textbf{Informer}}       & \scalebox{1.3}{78.69\std{1.68}} & \scalebox{1.3}{82.87\std{1.02}} & \scalebox{1.3}{69.19\std{2.90}} & \scalebox{1.3}{70.84\std{3.47}} & \scalebox{1.3}{92.09\std{0.53}} & \scalebox{1.3}{90.02\std{0.60}} & \scalebox{1.3}{80.62\std{1.70}} \\
    & \scalebox{1.2}{\textbf{iTransformer}}   & \scalebox{1.3}{\second{83.89\std{0.71}}} & \scalebox{1.3}{\second{88.25\std{1.18}}} & \scalebox{1.3}{76.39\std{1.01}} & \scalebox{1.3}{79.06\std{1.06}} & \scalebox{1.3}{91.18\std{1.16}} & \scalebox{1.3}{\second{90.93\std{0.98}}} & \scalebox{1.3}{\second{84.95\std{1.02}}} \\
    & \scalebox{1.2}{\textbf{MTST}}           & \scalebox{1.3}{76.59\std{1.90}} & \scalebox{1.3}{79.88\std{1.90}} & \scalebox{1.3}{66.31\std{2.95}} & \scalebox{1.3}{67.38\std{3.71}} & \scalebox{1.3}{86.86\std{2.75}} & \scalebox{1.3}{83.75\std{2.84}} & \scalebox{1.3}{76.80\std{2.68}} \\
    & \scalebox{1.2}{\textbf{Nonformer}}      & \scalebox{1.3}{78.66\std{0.49}} & \scalebox{1.3}{82.77\std{0.86}} & \scalebox{1.3}{69.12\std{0.87}} & \scalebox{1.3}{70.90\std{1.00}} & \scalebox{1.3}{89.37\std{2.51}} & \scalebox{1.3}{86.67\std{2.38}} & \scalebox{1.3}{79.58\std{1.35}} \\
    & \scalebox{1.2}{\textbf{PatchTST}}       & \scalebox{1.3}{74.74\std{1.62}} & \scalebox{1.3}{76.94\std{1.51}} & \scalebox{1.3}{63.89\std{2.71}} & \scalebox{1.3}{64.36\std{3.38}} & \scalebox{1.3}{88.79\std{0.91}} & \scalebox{1.3}{83.39\std{0.96}} & \scalebox{1.3}{75.35\std{1.85}} \\
    & \scalebox{1.2}{\textbf{Reformer}}       & \scalebox{1.3}{77.96\std{2.13}} & \scalebox{1.3}{81.72\std{1.61}} & \scalebox{1.3}{68.20\std{3.35}} & \scalebox{1.3}{69.65\std{3.88}} & \scalebox{1.3}{91.13\std{0.74}} & \scalebox{1.3}{88.42\std{1.30}} & \scalebox{1.3}{79.51\std{2.17}} \\
    & \scalebox{1.2}{\textbf{Transformer}}    & \scalebox{1.3}{77.37\std{1.02}} & \scalebox{1.3}{81.84\std{0.66}} & \scalebox{1.3}{67.14\std{1.80}} & \scalebox{1.3}{68.47\std{2.19}} & \scalebox{1.3}{90.08\std{1.76}} & \scalebox{1.3}{87.22\std{1.68}} & \scalebox{1.3}{78.69\std{1.52}} \\
    & \scalebox{1.2}{\textbf{Medformer}}      & \scalebox{1.3}{83.50\std{2.01}} & \scalebox{1.3}{85.19\std{0.94}} & \scalebox{1.3}{\second{77.11\std{3.39}}} & \scalebox{1.3}{\second{79.18\std{3.31}}} & \scalebox{1.3}{\second{92.81\std{1.48}}} & \scalebox{1.3}{90.32\std{1.54}} & \scalebox{1.3}{84.69\std{2.11}} \\
    
    \rowcolor{purple!22}& \scalebox{1.2}{\textbf{ViRe}}  & \scalebox{1.3}{\first{88.26\std{1.12}}} & \scalebox{1.3}{\first{89.54\std{0.78}}} & \scalebox{1.3}{\first{83.82\std{1.75}}} & \scalebox{1.3}{\first{85.81\std{1.52}}} & \scalebox{1.3}{\first{93.95\std{0.28}}} & \scalebox{1.3}{\first{93.08\std{0.72}}} & \scalebox{1.3}{\first{89.08\std{1.03}}}\\

    \midrule
    \multirow{11}{*}{\begin{tabular}[c]{@{}c@{}}\;\makecell{\scalebox{1.2}{\textbf{PTB-XL}} \\ \scalebox{1.2}{(5-Classes)}  \\ \scalebox{1.2}{(ECG)}} \end{tabular}} 
    & \scalebox{1.2}{\textbf{Autoformer}}     & \scalebox{1.3}{61.68\std{2.72}} & \scalebox{1.3}{51.60\std{1.64}} & \scalebox{1.3}{49.10\std{1.52}} & \scalebox{1.3}{48.85\std{2.27}} & \scalebox{1.3}{82.04\std{1.44}} & \scalebox{1.3}{51.93\std{1.71}} & \scalebox{1.3}{57.53\std{1.88}} \\
    & \scalebox{1.2}{\textbf{FEDformer}}      & \scalebox{1.3}{57.20\std{9.47}} & \scalebox{1.3}{52.38\std{6.09}} & \scalebox{1.3}{49.04\std{7.26}} & \scalebox{1.3}{47.89\std{8.44}} & \scalebox{1.3}{82.13\std{4.17}} & \scalebox{1.3}{52.31\std{7.03}} & \scalebox{1.3}{56.83\std{7.08}} \\
    & \scalebox{1.2}{\textbf{Informer}}       & \scalebox{1.3}{71.43\std{0.32}} & \scalebox{1.3}{62.64\std{0.60}} & \scalebox{1.3}{59.12\std{0.47}} & \scalebox{1.3}{60.44\std{0.43}} & \scalebox{1.3}{88.65\std{0.09}} & \scalebox{1.3}{64.76\std{0.17}} & \scalebox{1.3}{67.84\std{0.35}} \\
    & \scalebox{1.2}{\textbf{iTransformer}}   & \scalebox{1.3}{69.28\std{0.22}} & \scalebox{1.3}{59.59\std{0.45}} & \scalebox{1.3}{54.62\std{0.18}} & \scalebox{1.3}{56.20\std{0.19}} & \scalebox{1.3}{86.71\std{0.10}} & \scalebox{1.3}{60.27\std{0.21}} & \scalebox{1.3}{64.45\std{0.23}} \\
    & \scalebox{1.2}{\textbf{MTST}}           & \scalebox{1.3}{72.14\std{0.27}} & \scalebox{1.3}{63.84\std{0.72}} & \scalebox{1.3}{60.01\std{0.81}} & \scalebox{1.3}{61.43\std{0.38}} & \scalebox{1.3}{88.97\std{0.33}} & \scalebox{1.3}{65.83\std{0.51}} & \scalebox{1.3}{68.70\std{0.50}} \\
    & \scalebox{1.2}{\textbf{Nonformer}}      & \scalebox{1.3}{70.56\std{0.55}} & \scalebox{1.3}{61.57\std{0.66}} & \scalebox{1.3}{57.75\std{0.72}} & \scalebox{1.3}{59.10\std{0.66}} & \scalebox{1.3}{88.32\std{0.36}} & \scalebox{1.3}{63.40\std{0.79}} & \scalebox{1.3}{66.78\std{0.62}} \\
    & \scalebox{1.2}{\textbf{PatchTST}}       & \scalebox{1.3}{\first{73.23\std{0.25}}} & \scalebox{1.3}{\first{65.70\std{0.64}}} & \scalebox{1.3}{\first{60.82\std{0.76}}} & \scalebox{1.3}{\first{62.61\std{0.34}}} & \scalebox{1.3}{\first{89.74\std{0.19}}} & \scalebox{1.3}{\first{67.32\std{0.22}}} & \scalebox{1.3}{\first{69.90\std{0.40}}} \\
    & \scalebox{1.2}{\textbf{Reformer}}       & \scalebox{1.3}{71.72\std{0.43}} & \scalebox{1.3}{63.12\std{1.02}} & \scalebox{1.3}{59.20\std{0.75}} & \scalebox{1.3}{60.69\std{0.18}} & \scalebox{1.3}{88.80\std{0.24}} & \scalebox{1.3}{64.72\std{0.47}} & \scalebox{1.3}{68.04\std{0.52}} \\
    & \scalebox{1.2}{\textbf{Transformer}}    & \scalebox{1.3}{70.59\std{0.44}} & \scalebox{1.3}{61.57\std{0.65}} & \scalebox{1.3}{57.62\std{0.35}} & \scalebox{1.3}{59.05\std{0.25}} & \scalebox{1.3}{88.21\std{0.16}} & \scalebox{1.3}{63.36\std{0.29}} & \scalebox{1.3}{66.73\std{0.36}} \\
    
    & \scalebox{1.2}{\textbf{Medformer}}      & \scalebox{1.3}{72.87\std{0.23}} & \scalebox{1.3}{64.14\std{0.42}} & \scalebox{1.3}{\second{60.60\std{0.46}}} & \scalebox{1.3}{\second{62.02\std{0.37}}} & \scalebox{1.3}{\second{89.66\std{0.13}}} & \scalebox{1.3}{66.39\std{0.22}} & \scalebox{1.3}{69.28\std{0.31}} \\
    
    \rowcolor{purple!22}& \scalebox{1.2}{\textbf{ViRe}}  & \scalebox{1.3}{\second{73.12\std{0.24}}} & \scalebox{1.3}{\second{66.07\std{0.60}}} & \scalebox{1.3}{59.62\std{0.60}} & \scalebox{1.3}{61.59\std{0.47}} & \scalebox{1.3}{89.65\std{0.17}} & \scalebox{1.3}{\second{66.76\std{0.26}}} & \scalebox{1.3}{\second{69.47\std{0.39}}}\\

    \midrule
    \multirow{11}{*}{\begin{tabular}[c]{@{}c@{}}\;\makecell{\scalebox{1.2}{\textbf{MIMIC}} \\ \scalebox{1.2}{(2-Classes)}  \\ \scalebox{1.2}{(ECG)}} \end{tabular}} 
    & \scalebox{1.2}{\textbf{Autoformer}}     & \scalebox{1.3}{77.74\std{6.49}} & \scalebox{1.3}{77.92\std{6.53}} & \scalebox{1.3}{77.59\std{6.12}} & \scalebox{1.3}{77.58\std{6.37}} & \scalebox{1.3}{84.52\std{6.88}} & \scalebox{1.3}{82.77\std{7.19}} & \scalebox{1.3}{79.69\std{6.60}} \\
    & \scalebox{1.2}{\textbf{FEDformer}}      & \scalebox{1.3}{84.68\std{0.21}} & \scalebox{1.3}{84.63\std{0.28}} & \scalebox{1.3}{84.52\std{0.11}} & \scalebox{1.3}{84.56\std{0.18}} & \scalebox{1.3}{91.59\std{0.25}} & \scalebox{1.3}{90.78\std{0.24}} & \scalebox{1.3}{86.79\std{0.21}} \\
    & \scalebox{1.2}{\textbf{Informer}}       & \scalebox{1.3}{84.71\std{0.27}} & \scalebox{1.3}{84.60\std{0.22}} & \scalebox{1.3}{84.73\std{0.24}} & \scalebox{1.3}{84.65\std{0.22}} & \scalebox{1.3}{91.72\std{0.11}} & \scalebox{1.3}{91.23\std{0.19}} & \scalebox{1.3}{86.94\std{0.21}} \\
    & \scalebox{1.2}{\textbf{iTransformer}}   & \scalebox{1.3}{85.01\std{0.10}} & \scalebox{1.3}{84.94\std{0.14}} & \scalebox{1.3}{84.89\std{0.11}} & \scalebox{1.3}{84.91\std{0.16}} & \scalebox{1.3}{91.51\std{0.08}} & \scalebox{1.3}{91.37\std{0.12}} & \scalebox{1.3}{87.11\std{0.12}} \\
    & \scalebox{1.2}{\textbf{MTST}}           & \scalebox{1.3}{85.54\std{0.31}} & \scalebox{1.3}{85.44\std{0.32}} & \scalebox{1.3}{85.55\std{0.22}} & \scalebox{1.3}{85.48\std{0.38}} & \scalebox{1.3}{91.61\std{0.20}} & \scalebox{1.3}{91.49\std{0.14}} & \scalebox{1.3}{87.52\std{0.26}} \\
    & \scalebox{1.2}{\textbf{Nonformer}}      & \scalebox{1.3}{84.13\std{0.25}} & \scalebox{1.3}{84.03\std{0.17}} & \scalebox{1.3}{84.18\std{0.27}} & \scalebox{1.3}{84.08\std{0.16}} & \scalebox{1.3}{90.98\std{0.18}} & \scalebox{1.3}{90.72\std{0.21}} & \scalebox{1.3}{86.35\std{0.21}} \\
    & \scalebox{1.2}{\textbf{PatchTST}}       & \scalebox{1.3}{84.79\std{0.29}} & \scalebox{1.3}{84.71\std{0.31}} & \scalebox{1.3}{84.81\std{0.36}} & \scalebox{1.3}{84.73\std{0.32}} & \scalebox{1.3}{91.10\std{0.28}} & \scalebox{1.3}{90.61\std{0.32}} & \scalebox{1.3}{86.79\std{0.31}} \\
    & \scalebox{1.2}{\textbf{Reformer}}       & \scalebox{1.3}{\second{85.74\std{0.15}}} & \scalebox{1.3}{\second{85.75\std{0.18}}} & \scalebox{1.3}{\second{85.72\std{0.18}}} & \scalebox{1.3}{\second{85.88\std{0.18}}} & \scalebox{1.3}{\second{92.38\std{0.09}}} & \scalebox{1.3}{\second{91.94\std{0.19}}} & \scalebox{1.3}{\second{87.90\std{0.16}}} \\
    & \scalebox{1.2}{\textbf{Transformer}}    & \scalebox{1.3}{84.93\std{0.17}} & \scalebox{1.3}{84.84\std{0.16}} & \scalebox{1.3}{84.92\std{0.19}} & \scalebox{1.3}{84.86\std{0.16}} & \scalebox{1.3}{91.64\std{0.13}} & \scalebox{1.3}{90.89\std{0.16}} & \scalebox{1.3}{87.01\std{0.16}} \\
    & \scalebox{1.2}{\textbf{Medformer}}      & \scalebox{1.3}{85.10\std{0.33}} & \scalebox{1.3}{85.02\std{0.35}} & \scalebox{1.3}{85.12\std{0.38}} & \scalebox{1.3}{85.04\std{0.38}} & \scalebox{1.3}{91.44\std{0.23}} & \scalebox{1.3}{91.12\std{0.22}} & \scalebox{1.3}{87.14\std{0.32}} \\
    
    \rowcolor{purple!22}& \scalebox{1.2}{\textbf{ViRe}}  & \scalebox{1.3}{\first{88.61\std{0.12}}} & \scalebox{1.3}{\first{88.55\std{0.13}}} & \scalebox{1.3}{\first{88.56\std{0.15}}} & \scalebox{1.3}{\first{88.54\std{0.12}}} & \scalebox{1.3}{\first{95.08\std{0.06}}} & \scalebox{1.3}{\first{94.76\std{0.11}}} & \scalebox{1.3}{\first{90.68\std{0.12}}}\\

\bottomrule
\end{tabular}
}
\endgroup
\end{center}

\clearpage
\section{Computational Overhead Analysis}
\label{app:computational_overhead}
We measure APAVA F1-Score, per-batch latency, and peak memory at $B=128$. Training reuses cached CLIP features; online inference includes waveform rendering and vision encoding.
\begin{table}[!htb]
\centering
\caption{\textbf{Computational Overhead on APAVA.} F1-Score, per-batch latency ($B=128$), and peak memory are measured. Higher is better for F1-Score; lower is better for time and memory.}
\vspace{-3mm}
\label{tab:efficiency}
\small
\setlength{\tabcolsep}{8pt}
\renewcommand{\arraystretch}{1.08}
\resizebox{0.8\textwidth}{!}{%
\begin{tabular}{lccc}
\toprule
\textbf{Metric} & \textbf{ViRe} & \textbf{Medformer} & \textbf{FEDformer} \\
\midrule
F1-Score $\uparrow$ & \first{91.19\std{1.01}} & 76.31\std{0.71} & 73.51\std{3.30} \\
Training Time (ms) $\downarrow$ & \first{211.9} & 313.1 & 477.7 \\
Training Memory (MB) $\downarrow$ & \first{698.2} & 838.4 & 1103.9 \\
Inference Time (ms) $\downarrow$ & 267.7 & \first{152.6} & 282.6 \\
Inference Memory (MB) $\downarrow$ & 740.8 & \first{408.5} & 491.6 \\
\bottomrule
\end{tabular}%
}
\vspace{-2mm}
\end{table}

\paragraph{Training.} Caching removes repeated visual encoding from each optimization step. ViRe requires 211.9 ms and 698.2 MB per batch, compared with 313.1 ms and 838.4 MB for Medformer; the corresponding APAVA F1-Score is 91.19 versus 76.31 for the two models.

\paragraph{Inference.} The complete online path requires 267.7 ms and 740.8 MB. Its latency is lower than FEDformer (282.6 ms); the higher memory cost reflects the waveform rendering and CLIP vision encoding that are executed online in this measurement rather than read from the cache.

\paragraph{Normalized comparison.} Relative to Medformer, cached training lowers per-batch latency by 32.3\% and peak memory by 16.7\%, while raising the APAVA F1-Score by 14.88 points.

\paragraph{Deployment.} Cached features suit repeated optimization, whereas online use executes the full visual front end and is represented by the inference rows in \tabref{tab:efficiency}.

\paragraph{Trade-off.} Relative to Medformer, online inference is 1.75$\times$ slower and uses 1.81$\times$ more memory (267.7 vs. 152.6 ms; 740.8 vs. 408.5 MB), while the APAVA F1-Score improves by 14.88 points and training is cheaper. Whether the online cost is acceptable depends on the deployment budget, and caching removes most of it whenever the same recordings are scored repeatedly.

\paragraph{Measurement protocol.} All numbers are measured on the same NVIDIA RTX 4090 GPU used for the main experiments with batch size $B=128$. Training rows use cached CLIP features, whereas inference rows execute rendering and CLIP encoding online, so the cached and online settings respectively bound the practical training and deployment cost of ViRe from below and above.

\clearpage
\section{Interpretability Analysis of ViRe}
\label{app:interpretability}
\subsection{Qualitative Retrieval Attention Visualization}
\label{app:retrieval_attention}
\begin{figure}[!htb]
\centering
\makebox[\textwidth][c]{\includegraphics[width=\textwidth]{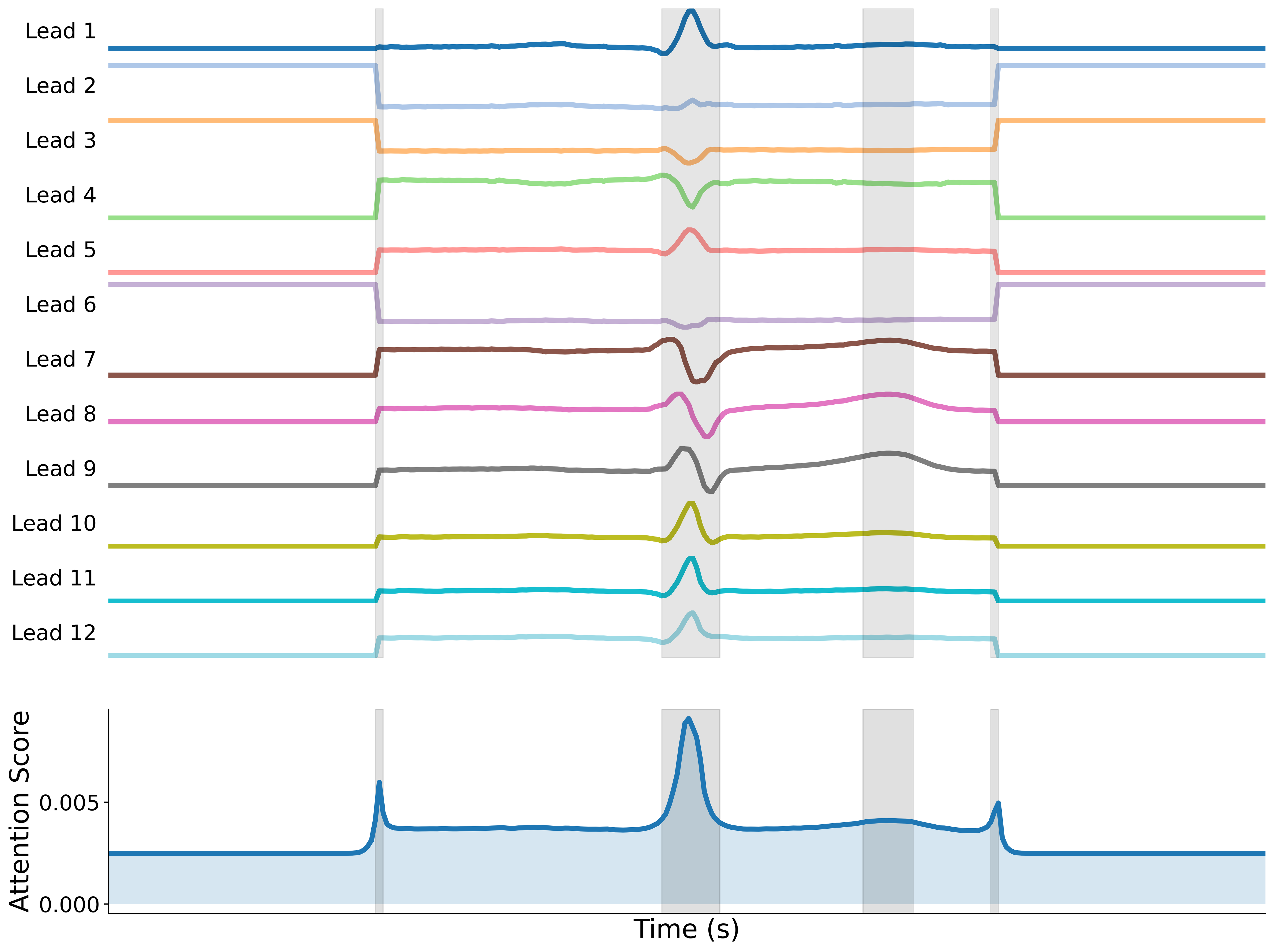}}
\vspace{-2mm}
\caption{\textbf{Retrieval Attention Visualization on PTB.} Temporal attention from the CLIP-derived Vision Query peaks around a morphologically salient interval with synchronous cross-lead variations.}
\label{fig:vision_retrieval_map}
\end{figure}

We inspect whether the CLIP-derived Vision Query retrieves physiologically meaningful temporal tokens. As shown in \figref{fig:vision_retrieval_map}, the temporal retrieval attention on PTB exhibits a localized peak around a morphologically salient interval, where multiple ECG leads show synchronous waveform changes. This pattern suggests that ViRe does not distribute attention uniformly over time; instead, the vision prior guides retrieval toward clinically relevant cross-lead variations.

\appref{app:attention_alignment} quantifies morphology and clinical alignment at the dataset level.

\subsection{Quantitative Attention Analysis}
\label{app:attention_alignment}
We further quantify whether retrieval attention is concentrated in meaningful temporal regions. We compare the CLIP-derived Vision Query with a non-semantic Gaussian Query baseline of the same dimensionality. Both metrics are attention-density ratios; values larger than one indicate denser attention inside the target region than in its temporal complement.

Let $X\in\mathbb{R}^{C\times T}$ be a channel-normalized ECG sample and let $A\in\mathbb{R}^{T}$ denote its temporal retrieval attention, with $A_t>0$ and $\sum_{t=1}^{T}A_t=1$. For a temporal region $\mathcal S$, define
\begin{equation}
\rho_A(\mathcal S)=\frac{1}{|\mathcal S|}\sum_{t\in\mathcal S}A_t.
\end{equation}
\eqparbreak
This length-normalized density makes short diagnostic intervals, such as the QRS region in ECG, comparable with their much longer complements in the same recording.

For morphology-aware alignment, we compute a channel-averaged second-order temporal variation score at every interior timestamp:
\begin{equation}
c_t=\frac{1}{C}\sum_{i=1}^{C}\left|X_{i,t+1}-2X_{i,t}+X_{i,t-1}\right|,\quad t=2,\ldots,T-1,
\end{equation}
and define $\mathcal S_{\mathrm{curv}}$ as the top-20\% timestamps ranked by $c_t$. For clinical alignment, $\mathcal S_{\mathrm{qrs}}$ denotes the timestamps covered by QRS segment(s), and $L_{\mathrm{qrs}}=|\mathcal S_{\mathrm{qrs}}|$. With complements denoted by bars, the Morphology-Aware Attention Ratio (MAR) and Clinical Alignment Ratio (CAR) are
\begin{equation}
\begin{aligned}
\mathrm{MAR}&=\frac{\rho_A(\mathcal S_{\mathrm{curv}})}{\rho_A(\bar{\mathcal S}_{\mathrm{curv}})},\\[-1mm]
\mathrm{CAR}&=\frac{\rho_A(\mathcal S_{\mathrm{qrs}})}{\rho_A(\bar{\mathcal S}_{\mathrm{qrs}})}.
\end{aligned}
\end{equation}
\eqparbreak
Thus, MAR evaluates enrichment on signal-intrinsic high-curvature morphology, whereas CAR evaluates enrichment on the clinically recognized QRS complex. \tabref{tab:attention_alignment} in the main text shows consistent ViRe gains over the Gaussian Query on both metrics and all three ECG datasets.

\clearpage
\section{Rendering Sensitivity Analysis}
\label{app:rendering_sensitivity}
\begin{figure}[!htb]
\centering
\includegraphics[width=0.95\textwidth]{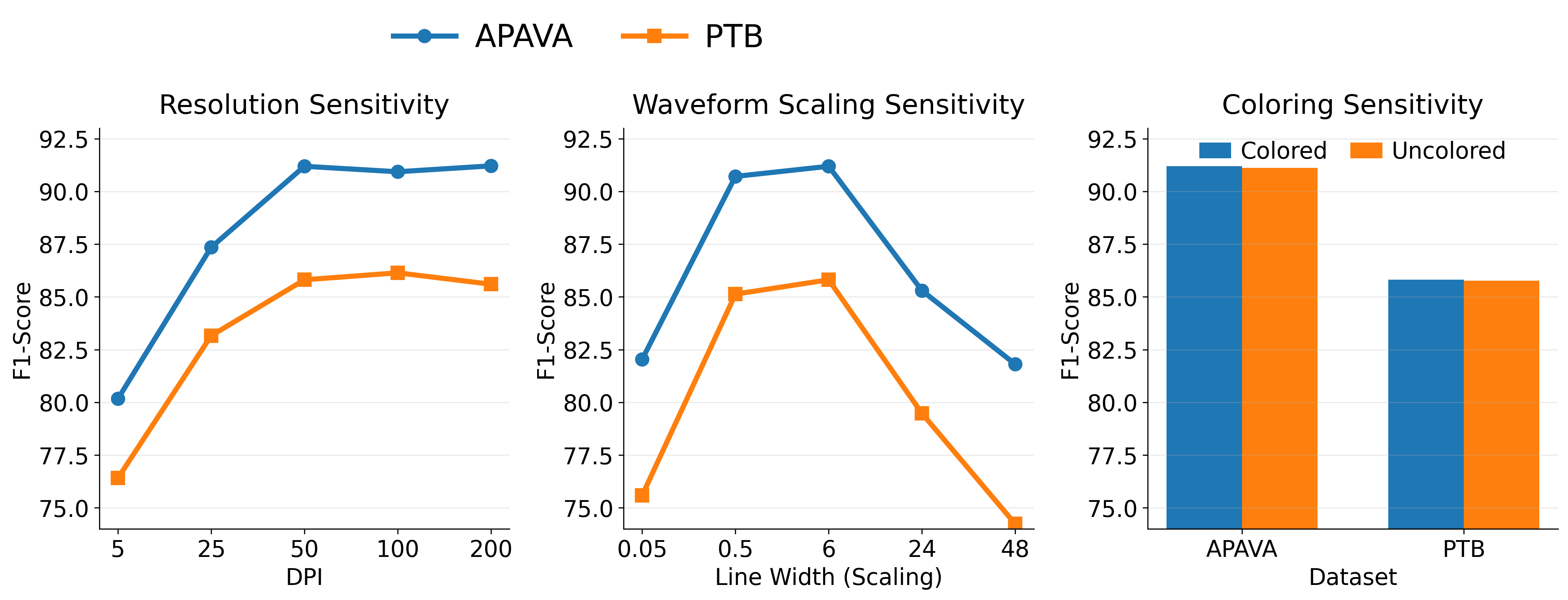}
\caption{\textbf{\textit{Rendering sensitivity of ViRe.}} F1-Score on APAVA and PTB under controlled variations of rendering resolution (left), waveform scaling/line width (middle), and channel layout coloring (right). The results show that ViRe is sensitive to excessively low resolution and inappropriate line width, while being largely insensitive to whether channels are colorized.}
\label{fig:rendering_sensitivity}
\end{figure}

We further study how the visualization operator depends on several rendering hyperparameters. Using the F1-Score from controlled ablations on APAVA and PTB, \figref{fig:rendering_sensitivity} summarizes the sensitivity of ViRe to \textbf{\textit{(i)}} rendering resolution (DPI), \textbf{\textit{(ii)}} waveform scaling (line width), and \textbf{\textit{(iii)}} channel layout (colored \textit{vs.} uncolored channels), with all other rendering settings fixed.

The figure reveals three consistent patterns. First, ViRe is clearly sensitive to extremely low rendering resolution: using only $5$ DPI causes a pronounced performance drop on both datasets, whereas moderate-to-high resolutions ($50$--$200$ DPI) are much more stable. Second, waveform scaling also matters: overly thin or overly thick lines degrade performance, suggesting that preserving an appropriate morphological thickness is important for CLIP-based visual encoding. Third, channel coloring has only a negligible influence, indicating that ViRe mainly benefits from the global waveform morphology rather than from color-specific channel cues.

\paragraph{Practical recommendation.} The results suggest rendering at moderate resolution ($50$--$200$ DPI) with a moderate line width, since both extremes degrade the CLIP representation, whereas channel coloring can be chosen freely. Rendering hyperparameters therefore deserve the same care as signal preprocessing when the operator is transferred to a new recording setup.

\clearpage
\section{Robustness and Transfer Analysis}
\label{app:robustness}
Beyond the primary benchmark, we evaluate subject-partition stability on repeated subject-disjoint splits and transfer to irregularly sampled medical time series.

\subsection*{Subject-Partition Stability}
Across held-out PTB-XL evaluations and repeated subject-disjoint splits, ViRe consistently improves over Medformer. \tabref{tab:stability} reports the paired statistical evidence for each evaluation.
\begin{table}[!htb]
\centering
\captionsetup{skip=2pt}
\caption{\textbf{Subject-partition stability and paired statistical evidence.}}
\label{tab:stability}
\small
\renewcommand{\arraystretch}{0.98}
\setlength{\tabcolsep}{3pt}
\begin{tabularx}{\textwidth}{@{} >{\raggedright\arraybackslash}p{0.42\textwidth} >{\centering\arraybackslash}p{0.06\textwidth} >{\centering\arraybackslash}p{0.06\textwidth} >{\raggedright\arraybackslash}X @{} }
\toprule
\rowcolor{softband}\textbf{Evaluation} & \textbf{Gain} & \textbf{Wins} & \textbf{Statistical evidence}\\
PTB-XL, 13 diagnoses (64.93 vs. 61.57) & \textbf{+3.36} & n/a & Cluster 95\% CI \textbf{[+1.46,+5.10]}; paired $p=0.0103$.\\
PTB-XL, five superclasses (72.26 vs. 68.52) & \textbf{+3.74} & n/a & Cluster 95\% CI \textbf{[+2.36,+5.14]}; paired $p=0.0043$.\\
\rowcolor{softband}PTB-XL, 13 diagnoses; repeated splits & \textbf{+2.46} & \textbf{7/7} & Wilcoxon $p=0.0156$; corrected paired $t$-test $p=0.0038$; 95\% CI \textbf{[+1.14,+3.77]}.\\
PTB-XL, five superclasses; repeated splits & \textbf{+2.42} & \textbf{7/7} & Wilcoxon $p=0.0156$; corrected paired $t$-test $p=0.0060$; 95\% CI \textbf{[+0.99,+3.84]}.\\
\rowcolor{softband}APAVA; repeated splits & \textbf{+6.48} & \textbf{8/10} & Exact Wilcoxon/Holm $p=0.0371$.\\
PTB; repeated splits & \textbf{+3.98} & \textbf{9/10} & Exact Wilcoxon/Holm $p=0.0078$.\\
\bottomrule
\end{tabularx}
\end{table}

\subsection*{Irregular Forecasting}
We follow the Hi-Patch protocol~\citep{luo2025hipatch} while retaining ViRe's visual-query retrieval mechanism. ViRe performs best on all six metrics and reduces MSE relative to Hi-Patch by 4.7\%, 6.2\%, and 10.9\% on Human Activity, PhysioNet, and MIMIC-III, respectively (\tabref{tab:irregular_forecasting}).
\begin{table}[!htb]
\centering
\captionsetup{skip=2pt}
\caption{\textbf{Irregular forecasting under the Hi-Patch protocol}~\citep{luo2025hipatch}. Lower is better.}
\label{tab:irregular_forecasting}
\small
\renewcommand{\arraystretch}{0.96}
\setlength{\tabcolsep}{2pt}
\begin{tabular*}{\textwidth}{@{\extracolsep{\fill}} l l r r r r @{}}
\toprule
\textbf{Dataset} & \textbf{Metric} & \textbf{ViRe} & \textbf{Hi-Patch} & \textbf{t-PatchGNN} & \textbf{GRU-D}\\
\midrule
\multirow{2}{*}{\textbf{Human Activity}} & MSE $\times10^{-3}$ & \textbf{2.45$\pm$0.04} & 2.57$\pm$0.02 & 2.66$\pm$0.03 & 3.94$\pm$0.29\\
& MAE $\times10^{-2}$ & \textbf{3.04$\pm$0.02} & 3.11$\pm$0.03 & 3.15$\pm$0.02 & 4.37$\pm$0.21\\
\midrule
\multirow{2}{*}{\textbf{PhysioNet}} & MSE $\times10^{-3}$ & \textbf{4.56$\pm$0.04} & 4.86$\pm$0.03 & 4.98$\pm$0.08 & 5.76$\pm$0.34\\
& MAE $\times10^{-2}$ & \textbf{3.44$\pm$0.03} & 3.62$\pm$0.07 & 3.72$\pm$0.03 & 4.53$\pm$0.15\\
\midrule
\multirow{2}{*}{\textbf{MIMIC-III}} & MSE $\times10^{-2}$ & \textbf{1.56$\pm$0.09} & 1.75$\pm$0.26 & 1.69$\pm$0.03 & 2.35$\pm$0.06\\
& MAE $\times10^{-2}$ & \textbf{6.77$\pm$0.11} & 7.24$\pm$0.18 & 7.22$\pm$0.09 & 8.34$\pm$0.22\\
\bottomrule
\end{tabular*}
\end{table}

\subsection*{Irregular Classification}
Following the MTM protocol, we compare ViRe with MTM, STraTS~\citep{tipirneni2022strats}, t-PatchGNN, and GRU-D~\citep{che2018grud} on P12, P19, and PAM. ViRe ranks first on all six reported metrics (\tabref{tab:irregular_classification}).
\begin{table}[!htb]
\centering
\captionsetup{skip=2pt}
\caption{\textbf{Irregular classification under the MTM protocol}~\citep{zhong2025mtm}. Higher is better.}
\label{tab:irregular_classification}
\small
\renewcommand{\arraystretch}{0.96}
\setlength{\tabcolsep}{2pt}
\begin{tabular*}{\textwidth}{@{\extracolsep{\fill}} l l r r r r r @{}}
\toprule
\textbf{Dataset} & \textbf{Metric} & \textbf{ViRe} & \textbf{MTM} & \textbf{STraTS} & \textbf{t-PatchGNN} & \textbf{GRU-D}\\
\midrule
\multirow{2}{*}{\textbf{P12}} & AUROC & \textbf{88.5$\pm$1.2} & 88.0$\pm$1.0 & 86.4$\pm$1.1 & 84.5$\pm$0.9 & 81.9$\pm$2.1\\
& AUPRC & \textbf{60.3$\pm$2.4} & 58.6$\pm$4.1 & 53.9$\pm$3.1 & 50.8$\pm$2.6 & 46.1$\pm$4.7\\
\midrule
\multirow{2}{*}{\textbf{P19}} & AUROC & \textbf{91.3$\pm$2.0} & 90.3$\pm$2.0 & 89.7$\pm$1.8 & 87.0$\pm$1.4 & 83.9$\pm$1.7\\
& AUPRC & \textbf{62.3$\pm$4.3} & 58.3$\pm$5.3 & 57.9$\pm$3.3 & 51.5$\pm$5.2 & 46.9$\pm$2.1\\
\midrule
\multirow{2}{*}{\textbf{PAM}} & Accuracy & \textbf{98.3$\pm$0.6} & 97.5$\pm$0.2 & 96.4$\pm$0.8 & 93.9$\pm$1.2 & 83.3$\pm$1.6\\
& F1 & \textbf{98.4$\pm$0.6} & 97.6$\pm$0.2 & 95.3$\pm$0.7 & 94.8$\pm$1.2 & 84.8$\pm$1.2\\
\bottomrule
\end{tabular*}
\end{table}

\clearpage
\section{Limitations and Societal Considerations}
\label{app:limitations_societal}

\paragraph{Limitations.}
ViRe is evaluated on six public EEG/ECG benchmarks with subject-independent splits, but retrospective results do not replace prospective multi-center clinical validation. Its visual prior is derived from a frozen CLIP vision encoder trained on general image-text data, which may miss clinically subtle waveform patterns that a domain-specific encoder could capture. The benchmark study covers EEG and ECG classification and, in \appref{app:robustness}, irregularly sampled forecasting and classification; other physiological signals and multi-label settings remain untested. Rendering hyperparameters influence the visual prior (\appref{app:rendering_sensitivity}), and online inference incurs the additional memory cost of on-the-fly rendering and CLIP encoding (\appref{app:computational_overhead}).

\paragraph{Societal considerations.}
ViRe is intended for decision support rather than autonomous diagnosis. It may improve data efficiency and morphology-aware EEG/ECG modeling, but risks remain, including over-reliance on automated outputs and privacy concerns in downstream use. All experiments use existing, de-identified datasets under their respective access terms, and no new patient data were collected. Clinical deployment, therefore, requires clinician oversight, privacy protection, and validation under the target distribution before any use in routine clinical practice.

\paragraph{Future directions.}
Promising extensions include vision-language encoders trained on medical waveform images, learned or adaptive rendering operators, and joint use of the frozen visual prior with textual reports, which resources such as MEETI make possible.

\newpage
\section*{NeurIPS Paper Checklist}

\begin{enumerate}

\item {\bf Claims}
    \item[] Question: Do the main claims made in the abstract and introduction accurately reflect the paper's contributions and scope?
    \item[] Answer: \answerYes{}
    \item[] Justification: The abstract and introduction state the waveform-morphology motivation, the ViRe framework, and the empirical scope; the quantitative claims are supported by Section~4, Table~\ref{tab:bench_summary}, and the conclusion.

\item {\bf Limitations}
    \item[] Question: Does the paper discuss the limitations of the work performed by the authors?
    \item[] Answer: \answerYes{}
    \item[] Justification: Limitations and potential deployment considerations are discussed in Appendix~\ref{app:limitations_societal}, including retrospective validation, reliance on a frozen CLIP vision encoder, and inference-time overhead.

\item {\bf Theory assumptions and proofs}
    \item[] Question: For each theoretical result, does the paper provide the full set of assumptions and a complete (and correct) proof?
    \item[] Answer: \answerYes{}
    \item[] Justification: The paper includes theoretical analysis in Appendix~\ref{app:vision}, where the setup, assumptions, Lemma~1, Theorem~1, Proposition~1, and their proofs are provided.

    \item {\bf Experimental result reproducibility}
    \item[] Question: Does the paper fully disclose all the information needed to reproduce the main experimental results of the paper to the extent that it affects the main claims and/or conclusions of the paper (regardless of whether the code and data are provided or not)?
    \item[] Answer: \answerYes{}
    \item[] Justification: The method, datasets, subject-independent splits, metrics, optimization protocol, and implementation details are described in Sections~3--4 and Appendices~\ref{app:data}--\ref{app:implementation}; code and training scripts are provided in the public repository at \url{https://github.com/Levi-Ackman/ViRe}.

\item {\bf Open access to data and code}
    \item[] Question: Does the paper provide open access to the data and code, with sufficient instructions to faithfully reproduce the main experimental results, as described in supplemental material?
    \item[] Answer: \answerYes{}
    \item[] Justification: The code and training scripts are released at \url{https://github.com/Levi-Ackman/ViRe}, and public dataset sources are cited with access links.

\item {\bf Experimental setting/details}
    \item[] Question: Does the paper specify all the training and test details (e.g., data splits, hyperparameters, how they were chosen, type of optimizer) necessary to understand the results?
    \item[] Answer: \answerYes{}
    \item[] Justification: Section~4 specifies the evaluation metrics, random seeds, hardware, and baseline protocol, while Appendices~\ref{app:data} and~\ref{app:implementation} provide dataset preprocessing, subject-level splits, optimizer, batch sizes, early stopping, and hyperparameter settings.

\item {\bf Experiment statistical significance}
    \item[] Question: Does the paper report error bars suitably and correctly defined or other appropriate information about the statistical significance of the experiments?
    \item[] Answer: \answerYes{}
    \item[] Justification: All main benchmark and ablation tables report mean and standard deviation over five random seeds; Section~4 states the reporting protocol and Table~\ref{tab:full_results} reports metric-wise mean$\pm$std values.

\item {\bf Experiments compute resources}
    \item[] Question: For each experiment, does the paper provide sufficient information on the computer resources (type of compute workers, memory, time of execution) needed to reproduce the experiments?
    \item[] Answer: \answerYes{}
    \item[] Justification: Section~4 reports the GPU type used for experiments, Appendix~\ref{app:computational_overhead} reports per-batch latency and peak memory, and Appendix~\ref{app:implementation} gives batch sizes and implementation settings.
    
\item {\bf Code of ethics}
    \item[] Question: Does the research conducted in the paper conform, in every respect, with the NeurIPS Code of Ethics \url{https://neurips.cc/public/EthicsGuidelines}?
    \item[] Answer: \answerYes{}
    \item[] Justification: The work uses cited EEG/ECG datasets under their official access protocols and is presented as a decision-support research method rather than an autonomous clinical diagnostic system, consistent with the NeurIPS Code of Ethics.

\item {\bf Broader impacts}
    \item[] Question: Does the paper discuss both potential positive societal impacts and negative societal impacts of the work performed?
    \item[] Answer: \answerYes{}
    \item[] Justification: Appendix~\ref{app:limitations_societal} discusses positive impacts, such as morphology-aware medical time-series modeling, and negative risks, including erroneous predictions, over-reliance, and privacy concerns.
    
\item {\bf Safeguards}
    \item[] Question: Does the paper describe safeguards that have been put in place for responsible release of data or models that have a high risk for misuse (e.g., pre-trained language models, image generators, or scraped datasets)?
    \item[] Answer: \answerNA{}
    \item[] Justification: The paper does not release high-risk generative models, scraped datasets, or patient-level data; it releases code/training scripts for reproducing the proposed classification framework.

\item {\bf Licenses for existing assets}
    \item[] Question: Are the creators or original owners of assets (e.g., code, data, models), used in the paper, properly credited and are the license and terms of use explicitly mentioned and properly respected?
    \item[] Answer: \answerYes{}
    \item[] Justification: Existing datasets, baselines, and the CLIP vision encoder are credited through citations and URLs in Sections~3--4 and Appendices~\ref{app:data}--\ref{app:implementation}; the paper does not redistribute the original datasets.

\item {\bf New assets}
    \item[] Question: Are new assets introduced in the paper well documented and is the documentation provided alongside the assets?
    \item[] Answer: \answerYes{}
    \item[] Justification: The new asset is the implementation and training scripts for ViRe, documented by the method description, the implementation appendix, and the public repository at \url{https://github.com/Levi-Ackman/ViRe}.

\item {\bf Crowdsourcing and research with human subjects}
    \item[] Question: For crowdsourcing experiments and research with human subjects, does the paper include the full text of instructions given to participants and screenshots, if applicable, as well as details about compensation (if any)? 
    \item[] Answer: \answerNA{}
    \item[] Justification: The paper does not conduct crowdsourcing experiments or collect new human-subject data; it uses existing EEG/ECG benchmark datasets.

\item {\bf Institutional review board (IRB) approvals or equivalent for research with human subjects}
    \item[] Question: Does the paper describe potential risks incurred by study participants, whether such risks were disclosed to the subjects, and whether Institutional Review Board (IRB) approvals (or an equivalent approval/review based on the requirements of your country or institution) were obtained?
    \item[] Answer: \answerNA{}
    \item[] Justification: The paper does not collect new human-subject data. It uses existing datasets under their original access and preprocessing protocols, so new IRB approval by the authors is not applicable.

\item {\bf Declaration of LLM usage}
    \item[] Question: Does the paper describe the usage of LLMs if it is an important, original, or non-standard component of the core methods in this research? Note that if the LLM is used only for writing, editing, or formatting purposes and does \emph{not} impact the core methodology, scientific rigor, or originality of the research, declaration is not required.
    \item[] Answer: \answerNA{}
    \item[] Justification: LLMs are not used as an important or non-standard component of the proposed method. The method uses a frozen CLIP vision encoder as a visual feature extractor, which is described in Section~3.

\end{enumerate}
\end{document}